\documentclass[10pt,journal,compsoc]{IEEEtran}

\IEEEoverridecommandlockouts
\usepackage{cite}
\usepackage{amsmath,amssymb,amsfonts}
\usepackage{algorithmic}
\usepackage{graphicx}
\usepackage{textcomp}
\usepackage{xcolor}
\def\BibTeX{{\rm B\kern-.05em{\sc i\kern-.025em b}\kern-.08em
    T\kern-.1667em\lower.7ex\hbox{E}\kern-.125emX}}

\usepackage{hyperref}
\usepackage{url}
\usepackage{multirow}
\usepackage{caption}
\usepackage{subcaption}
\usepackage{graphicx}
\usepackage{amsmath}
\usepackage{wrapfig}
\usepackage{comment}

\usepackage{amsmath,amsfonts,bm}

\usepackage{colortbl}

\usepackage{enumitem,algorithm,algorithmic}
\usepackage{threeparttable}
\usepackage{tabularx}
\usepackage{booktabs}
\usepackage{algorithm}
\usepackage{algorithmic}
\newtheorem{proposition}{Proposition}
\newtheorem{proof}{Proof}

\def\d{\mathrm{d}}

\usepackage{natbib}

\setcitestyle{numbers,square}

\begin{document}

\title{Improving Subgraph Extraction for Graph Invariant Learning via Graph Sinkhorn Attention
}

\author{
        Junchi Yan,~\textit{Senior Member, IEEE},
        Fangyu Ding,
        Jiawei Sun, 
        Zhaoping Hu,
        Yunyi Zhou,
        %Haiyang Wang,
        and Lei Zhu
        % Zhixuan Chu,
        %and Tianming Li
% School of Artificial Intelligence
\IEEEcompsocitemizethanks{\IEEEcompsocthanksitem J. Yan, F. Ding, J. Sun and Z. Hu are with School of Artificial Intelligence and School of Computer Science, Shanghai Jiao Tong University, Shanghai 200240, China. E-mail:
\{yanjunchi,noelsjw,zhaopinghu\}@sjtu.edu.cn, fangyu.ding@outlook.com. Correspondence authors: F. Ding and J. Sun.}

\IEEEcompsocitemizethanks{\IEEEcompsocthanksitem Y. Zhou, L. Zhu are with Ant Group, Hangzhou 310023, China. (E-mail:
zhouyunyi.zyy, simon.zl\}@antgroup.com).}
}
%, william.ltm
%chuzhixuan.czx,
%% we raise the question: \textit{How to Improve Graph Invariant Learning via Subgraph Extraction?} 
% To answer it, 
\maketitle

%Despite their success, such methods also have various limitations in obtaining the invariant subgraphs.
% Graph invariant learning (GIL) has been a popular approach, which in general aims to discovering the invariant relationships among graphs and their labels, under various distribution shifts. One step further, extensive endeavors have been recently devoted to extracting the invariant subgraph from the input graph to achieve generalization. 
\begin{abstract}
Graph invariant learning (GIL) seeks invariant relations between graphs and labels under distribution shifts. Recent works try to extract an invariant subgraph to improve out-of-distribution (OOD) generalization, yet existing approaches either lack explicit control over compactness or rely on hard top-$k$ selection that shrinks the solution space and is only partially differentiable. In this paper, we provide an in-depth analysis of the drawbacks of some existing works and propose a few general principles for invariant subgraph extraction: 1) separability, as encouraged by our sparsity-driven mechanism, to filter out the irrelevant common features; 2) softness, for a broader solution space; and 3)  differentiability, for a soundly end-to-end optimization pipeline. Specifically, building on optimal transport, we propose Graph Sinkhorn Attention (GSINA), a fully differentiable, cardinality-constrained attention mechanism that assigns sparse-yet-soft edge weights via Sinkhorn iterations and induces node attention. GSINA provides explicit controls for separability and softness, and uses a Gumbel reparameterization to stabilize training. It convergence behavior is also theoretically studied. Extensive empirical experimental results on both synthetic and real-world datasets validate its superiority.
\end{abstract}
% To jointly encourage these principles, we specifically leverage the Optimal Transport (OT) framework, and propose a graph attention mechanism called Graph Sinkhorn Attention (GSINA). It serves as a regularization method for GIL tasks, with the hope of obtaining meaningful invariant subgraphs, in a differentiable way with controllable sparsity and softness.

%Moreover, GSINA is a general graph learning framework that could handle GIL tasks of multiple data grain levels. 


\section{Introduction}\label{sec:intro}

% 1. Graph, GNN, Graph OOD
% 2. GIL
% 3. analysis
% 4. ours
% contributions

Graph is ubiquitous in real world, e.g. social networks~\cite{zhang2022improving}, supply chain networks~\cite{yang2021financial}, and chemical molecules etc. Graph machine learning, especially by the architecture of graph neural networks (GNNs)~\cite{xu2018powerful}, has shown promising results in various graph-related tasks~\cite{GAT,sui2022causal}. Despite their empirical success, many existing approaches still often rely on the Independent and Identically Distributed (I.I.D.) assumption, assuming the training and testing graph data are drawn from the same distribution. However, distribution shifts commonly occur in complex graph data, making out-of-distribution (OOD) generalization~\cite{yang2022learning} a crucial problem in practical graph learning.

% distribution shifts, i.e., the mismatches over training and testing domains widely exist especially for complex graph data. The out-of-distribution (OOD) generalization~\cite{yang2022learning} has become a heated topic in recent graph representation learning.

% the GIL research lines 
As an effective way to achieve graph OOD generalization, Graph invariant learning (GIL), which aims to capture the invariant relationships between graph data and labels, has been extensively studied in different settings including both graph-level~\cite{li2022learning,buffelli2022sizeshiftreg,wu2022discovering} and node-level~\cite{2019Disentangled,wu2022handling} tasks. Existing GIL methods can be broadly categorized into two types: explicit representation alignment and invariance optimization. The former aligns graph representations across multiple environments by minimizing their distributional discrepancies via regularization~\cite{buffelli2022sizeshiftreg,wu2022handling}. The latter builds on the invariance principle, seeking invariant features or properties within data under distribution shifts~\cite{wu2022discovering,miao2022interpretable,chen2022learning}.
Empirically, key graph information often resides in a few crucial nodes or edges—for example, functional groups determining molecular solubility~\cite{wencel2013c}, or core members influencing community risk in financial networks~\cite{wang2019semi}.

% GIL can be roughly categorized into two lines of research: namely explicit representation alignment and invariance optimization. The general idea of the explicit representation alignment methods is to align graph representations among multiple environments, and they are often designed to minimize the difference across environments with certain regularization strategies~\cite{buffelli2022sizeshiftreg,wu2022handling}. On the other hand, the invariance optimization methods, are based on the principle of invariance, which assume the invariant property inside data or the invariant features under distribution shifts. 
% Many of these methods are aimed at handling graph OOD generalization by discovering graph invariant features (e.g. crucial nodes and edges) under distribution shifts~\cite{wu2022discovering,miao2022interpretable,chen2022learning}. 
% As empirical collateral evidence, crucial graph information usually exists in a few edges and nodes in real-world scenarios. E.g., in chemistry, key functional groups in a molecule yield a certain property like solubility~\cite{wencel2013c}. For finance, the risk level of a community is often mainly determined by a few key members~\cite{wang2019semi}. 

% state the problems of previous work

% 已有方法在做什么，以及问题（sparse soft的缺乏）
% GSAT问题：用IB约束子图，sparsity
% TopK问题：TopK算子partially differential，子图是card
% card问题非零个数小于等于k，soft card等于k

% 引出card，引出问题，为什么需要，soft-card问题为什么提出SCardGIL
% 参考 摘要

\begin{figure*}[tb!]
     \centering
     \begin{subfigure}[htb]{0.4\textwidth}
         \centering
         \includegraphics[width=\textwidth]{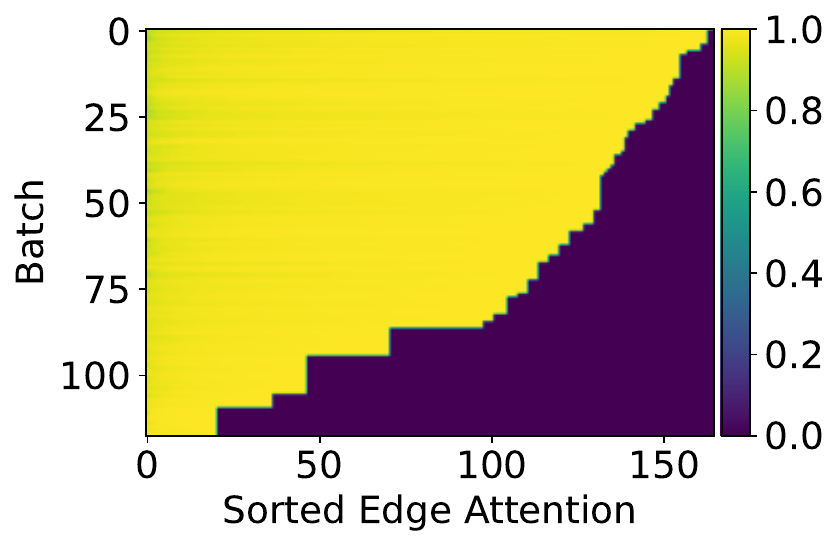}
         %%\vspace{-20pt}
         \caption{Heatmap (GSAT)}
         \label{fig:gsat_demo}
     \end{subfigure}
     \begin{subfigure}[htb]{0.4\textwidth}
         \centering
         \includegraphics[width=\textwidth]{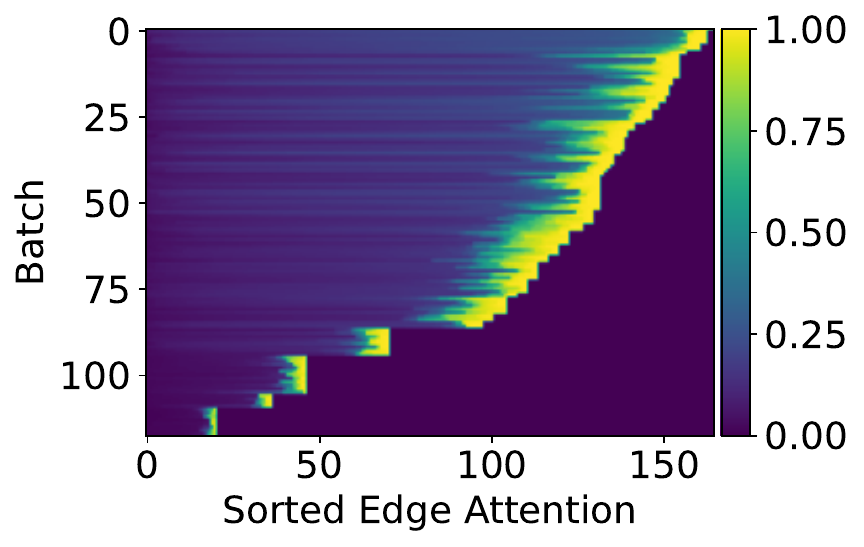}
         %%\vspace{-20pt}
         \caption{Heatmap (proposed GSINA)}
         \label{fig:ours_demo}
     \end{subfigure}
     \begin{subfigure}[htb]{0.39\textwidth}
         \centering
         \includegraphics[width=\textwidth]{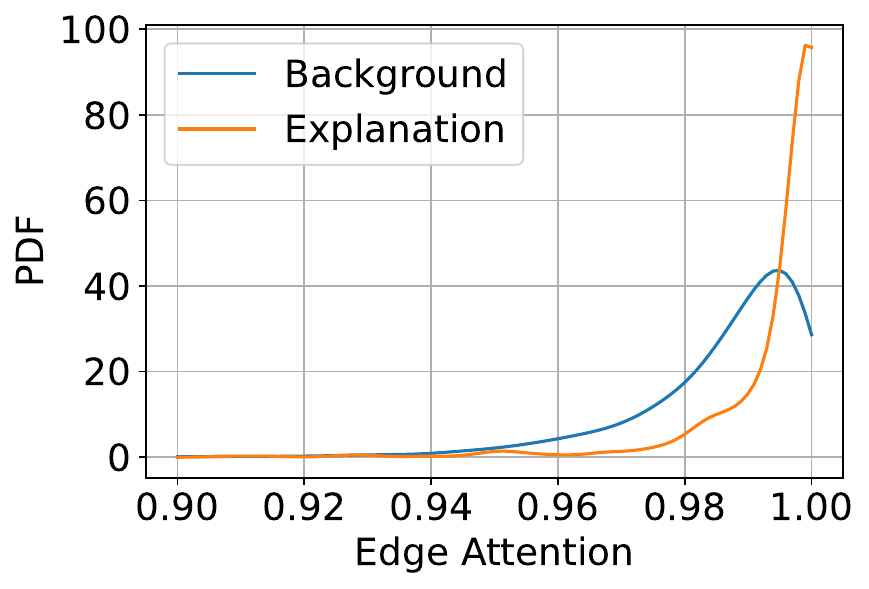}
         %%\vspace{-20pt}
         \caption{PDF (GSAT)}
         \label{fig:gsat_pdf}
     \end{subfigure}
     \begin{subfigure}[htb]{0.37\textwidth}
         \centering
         \includegraphics[width=\textwidth]{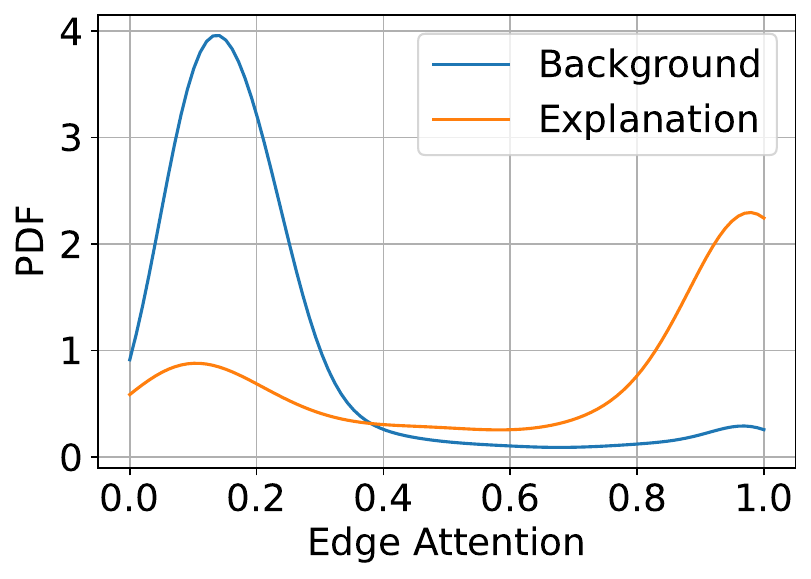}
         %%\vspace{-20pt}
         \caption{PDF (proposed GSINA)}
         \label{fig:ours_pdf}
     \end{subfigure}
         %%\vspace{-5pt}
     \caption{Subgraph separability of GSAT~\cite{miao2022interpretable} and our proposed GSINA (with subgraph extraction ratio $r=0.3$) on a batch of samples in SPMotif (with graph generation hyperparamter $b=0.5$) dataset (from DIR~\cite{wu2022discovering}). Note that Fig.~\ref{fig:gsat_demo} and Fig.~\ref{fig:ours_demo} demonstrate the learned edge importance (i.e. attention values) by GSAT and GSINA for each edge of the graphs in the batch, and the bottom right black regions are non-edge padding. The X-axis represents edges (sorted by attention values), and the Y-axis indexes the graphs. Note that here the graph sizes vary. For visualization we pad each row to the batch maximum edge count; the black cells on the right/bottom indicate padding (non-edges). The color intensity reflects the relative magnitude of attention scores across the edges, not necessarily their final normalized values. Also note that for each graph and each method independently, we sort that graph’s edge-attention scores in descending order and place them left→right within that row. The x-axis is therefore edge rank (sorted within graph; 1 = largest), the y-axis is graph ID, and the color encodes the (unnormalized) attention value. Because sorting is per-method, the per-row order in (a) and (b) need not match; small apparent inversions can arise from ties and color quantization after padding. Fig.~\ref{fig:gsat_pdf} and Fig.~\ref{fig:ours_pdf} are the PDF (probability density function) of edge attention generated by GSAT and GSINA for the edges in background (label-independent) part and explanation (label-related) part. 
     % Both GSAT and our GSINA are trained to converge with early stopping patience $10$.
     }
         %%\vspace{-10pt}
\label{fig:demo}
\end{figure*}

In particular, the invariance optimization methods~\cite{miao2022interpretable,wu2022discovering,chen2022learning} have made considerable efforts for sound generalizability and inherent interpretability of their invariant feature discovery, which bear two mainstream schemes: 1) the \emph{Information Bottleneck (IB)}~\cite{tishby2015deep}   scheme~\cite{yu2020graph} which exploits the IB principle to extract label-relevant graph invariant features with information constraint; 2) the \emph{top-$k$ Subgraph Selection} scheme~\cite{wu2022discovering,chen2022learning} that intuitively chooses the top-$k$ most influential edges as the `invariant subgraph'. Despite these advances, several limitations remain.
% Despite the advance made by those works, there are still limitations that need to be stated and addressed. 
Specifically, for the IB based methods, they might not proactively guarantee the separability of subgraphs, as also mentioned in~\cite{miao2022interpretable}.  Fig.~\ref{fig:demo} shows the behavior of the seminal IB based GIL method: Graph Stochastic Attention (GSAT)~\cite{miao2022interpretable}, which evaluates the importance of each edge by assigning edge attention. Specifically, as shown in Fig.~\ref{fig:gsat_demo} and Fig.~\ref{fig:gsat_pdf}, 
% compared with ours
GSAT lacks separability: the edge attention for the background part  (label-independent) and the explanation part  (label-dependent) are similar, making it difficult to make a prediction based on the most valuable invariant features, which are supposed to be more distinguishable. 

One may argue that the above plot results are not absolutely sparse. It is worth noting that, our intent is not to claim GSAT never produces near-zero scores, but that GSAT lacks an explicit, controllable cardinality constraint and often yields less separation between background and explanatory edges. To avoid confusion, we will use ``separability (distinguishability)” when referring to Fig. 1, and reserve ``sparsity” for our method’s explicit ratio control $r$ in GSINA. The PDFs in Fig. 1c–1d precisely capture this point: GSAT’s two distributions overlap substantially, whereas GSINA’s are well separate.

% Regarding the reviewer's concern about the uniform high values in Fig. 1a: GSAT uses a stochastic sampling process for selecting top-$k$ edges based on learned logits (without softmax), and these values tend to saturate towards the upper bound (often close to 1) for selected edges and near 0 for others due to the sharpening effect of the hard top-$k$ selection mechanism. This explains why the GSAT heatmap shows large contiguous blocks of bright yellow—these indicate that a fixed number of edges (e.g., top 30\%) consistently receive near-maximum values, creating a hard sparsity pattern. In contrast, GSINA applies a soft and differentiable attention mechanism (via Sinkhorn), resulting in smoother value transitions, as shown in Fig. 1b.

% To clarify, the values in the heatmaps are not post-softmax outputs. In GSINA, edge scores are regularized via a differentiable Sinkhorn transport plan, while GSAT produces logits that are then used for sampling but are not passed through a softmax. Hence, it's not necessarily expected that the GSAT attention values sum to 1 or resemble typical soft attention distributions.

For the top-$k$ based methods, they try to capture the invariant subgraph in a `hard' way, i.e. only the top-$k$ part is kept for training and prediction, and the other part is neglected. This hard subgraph extraction restricts the available information and thus narrows the solution space for discovering the optimal invariant subgraph. Moreover, the top-$k$ operation is inherently non-differentiable, posing an ill-posed optimization problem since it prevents gradient backpropagation. To enable training, recent works~\cite{wu2022discovering,chen2022learning} introduce partial differentiability by assigning continuous weighting scores from the subgraph extractor to the selected top-$k$ subgraph. However, because the top-$k$ selection removes part of the graph structure, these methods are limited to graph-level tasks and are inapplicable to fine-grained node-level scenarios where discarded nodes lead to incomplete graph representations.
% As only restricted information is used, hard subgraph extraction leads to a restricted solution space to obtain the optimal invariant subgraph. It involves an ill-posed  problem: the top-$k$ selection operation itself is not differentiable (it does not provide gradients for model backward pass). To make them `trainable', these methods are partially differentiable for learning after assigning the differentiable weighting scores output from their subgraph extractors for \emph{top-$k$} selection to the extracted subgraph (see~\cite{wu2022discovering,chen2022learning}). Moreover, as the \emph{top-$k$} selection discards part of the graph structure, these methods could only be used at graph level. While for fine-grained node level tasks, they are not applicable as the graph structure is incomplete, as part of the nodes have been discarded. 

%There lacks a way to learn all the node representations.

\begin{table}[tb!]%{R}{0.45\columnwidth}
% %%\vspace{-5pt}
\caption{Comparison of GIL methods based on subgraph learning, and our method achieves all the three properties. } % GIB不是GIL方法
%%\vspace{-10pt}
%Specifically, as will be shown later in the paper, we use the hyperparameter: sparsity ratio $r$ and softness $\tau$ to control our algorithm.
\begin{center}
% \begin{small}
% \resizebox{0.45\columnwidth}{!}{
\begin{tabular}{lccc}
\toprule
  & Sparse                  & Soft                & Differentiable                  \\
\midrule
 GSAT~\cite{miao2022interpretable} (IB) & No & Yes  & Fully \\
 % DIR~\cite{wu2022discovering} (top-$k$) & Yes & No & Partially \\
 CIGA~\cite{chen2022learning} (top-$k$) & Yes & No & Partially \\
 
 GSINA (ours) & \textbf{Yes} & \textbf{Yes} &  \textbf{Fully}\\
\bottomrule
\end{tabular}
% }
\end{center}
\label{tab:methods}
     %%\vspace{-25pt}
\end{table}

% introduce our method on the whole
% 参考sec 3
%\textit{How to extract the invariant subgraph for graph invariant learning?} 
Specifically, we summarize the three desirable properties for an invariant subgraph extractor to address the above-mentioned issues of invariance optimization GIL: 
1) separability (see Fig.~\ref{fig:ours_demo},~\ref{fig:ours_pdf}) to effectively filter out the variant features, and the invariant subgraph should idealy be sufficiently distinguishable to avoid confusion with the variant part. In the following part of this paper, we use ``separability" as the objective and treat ``sparsity" as a controllable means to promote separability via sparse node attention with aligned edge weights~\cite{blandford2003compact,quesada2018separable}. 2) softness (compared with the `hard selection' ways) to enlarge the subgraph solution space, to numerically evaluate the graph feature importance by maintaining graph information; and 3) differentiability (also based on softness) for soundly end-to-end optimization, and in turn it ensures the invariant subgraph extractor could be learned to generate separable and soft subgraphs.

To achieve these properties, we have also taken inspiration from the designs and characteristics of existing GIL works~\cite{wu2022discovering,miao2022interpretable,chen2022learning}, as summarized in Table~\ref{tab:methods}. Specifically, GSAT~\cite{miao2022interpretable} is soft and differentiable as it is based on graph attention mechanism. While DIR~\cite{wu2022discovering} and CIGA~\cite{chen2022learning} are sparse as their top-$k$ operations explicitly constrain the proportion of their subgraphs to their input graphs. Additionally, recent advance in cardinality-constrained combinatorial optimization~\cite{wang2023towards} shows that the top-$k$ problem could be addressed in an approximated soft version by a series of soft and differentiable iterative numerical calculations of the Optimal Transport (OT)~\cite{peyre2019computational} by using the Sinkhorn algorithm~\cite{Sinkhorn1964ARB}, with bounded constraint violations. Building upon these insights, we propose a novel and general graph attention mechanism, termed Graph Sinkhorn Attention (GSINA), for enhancing GIL tasks across multiple levels. GSINA evaluates the importance of graph features (nodes and edges) by assigning sparse yet soft attention values.
% Based on these fundamental endeavors, we propose a novel and general graph attention mechanism, which we call it Graph Sinkhorn Attention (GSINA) for improving GIL tasks of multiple levels, which evaluates the graph features (including nodes and edges) importance by assigning sparse and soft graph attention values.
As an invariance optimization method, GSINA defines its invariant subgraph in the manner of graph attention, which serves as a powerful regularization. The highlights of the paper are:
% 通用框架
\begin{itemize}
    \item We analyze and formalize the usefulness of separability (encouraged by sparsity), softness, and differentiability in invariant subgraph extraction for Graph Invariant Learning (GIL), aspects that remain underexplored in previous IB and top-$k$ based methods.
    % We summarize and analyze the necessity of sparsity, softness, and differentiability in subgraph extraction for graph invariance learning (GIL), which has not been well studied in the literature of previous IB and top-$k$ based methods.
    \item We propose Graph Sinkhorn Attention (GSINA), a GIL framework that learns fully differentiable invariant subgraphs with a controllable scheme on separability and softness, enabling generalization across tasks.
    \item We theoretically show the exponential convergence speed of our adopted methods.
    % We propose Graph Sinkhorn Attention (GSINA), a GIL framework by learning fully differentiable invariant subgraphs with controllable sparsity and softness to improve graph generalization tasks at different levels. 
    \item Experiments on public benchmarks show its effectiveness, including both synthetic and real-world datasets.
\end{itemize}

%The paper is organized as follows. Sec.~\ref{sec:back} discusses the related work. Sec.~\ref{sec:method} presents our main methods. Experimental results and discussion are given in Sec.~\ref{sec:exp} and Sec.~\ref{sec:con} concludes the paper.

\section{Preliminaries and Related Works}\label{sec:back}
% 
% \subsection{Problem Formulation and Related Work}
The related works cover different aspects related to our work, including the problem formulation of graph out-of-distribution (OOD) generalization and Graph Invariant Learning (GIL), the inductive bias behind invariant subgraph extraction, and the cardinality-constrained combinatorial optimization for a fully differentiable top-$k$ operation.

%\yjc{need to add more up-to-date references!}
%%\vspace{-3pt}

\subsection{Graph OOD Generalization}

%%\vspace{-3pt}
% 
 % Our work involves a set of graph datasets  $\dataset=\{\dataset^e\}_e$ from multiple domains $\mathcal{E}_{\text{all}}$. Samples $(G^e_i, Y^e_i)\in \dataset^e$ from the same domain are considered independently drawn from an identical distribution. A general framework denoted $\rho \circ h$ involves an encoder $h:\mathcal{G} \rightarrow \mathbb{R}^h$ to learn a representation $\mathbf{h}_G$ for each graph $G$ and a function $\rho:\mathbb{R}^h \rightarrow \mathcal{Y}$  that aids in predicting the label $\hat{Y}_G=\rho(\mathbf{h}_G)$ for a downstream task. The goal is to train the $\rho \circ h$ from training domains $\mathcal{D_{\mathrm{tr}}}=\{\dataset^e\}_{e\in\mathcal{E}_{\text{tr}}\subseteq\mathcal{E}_{\text{all}}}$ such that it generalizes well to unseen domains, i.e., we aim to minimize $\max_{e\in\mathcal{E}_{\text{all}}}R^e$, where $R^e$ represents the empirical risk of  $\rho\circ h$ under domain $e$.

%  \subsection{Problem Formulation Details}
% Here we provide the problem formulation details of our GSINA, as well as discussions of graph out-of-distribution (OOD) settings. 

% We consider a graph dataset of $\mathcal{G} = \{(G_i, Y_i)\}_{i=1}^{N}$
% \textbf{Graph OOD Generalization.} 
In this paper, the concept of invariant learning specifically refers to utilizing the invariant relationships between features and labels across various distributions while disregarding any spurious correlations that may arise~\cite{li2022learning}. In this way, it is possible to attain a high level of out-of-distribution (OOD) generalization in the presence of distribution shifts, which has become an active area in recent years ~\cite{OODSurvey2024}.

For the graph OOD problem considered here, we assume a set of graph datasets $\mathcal{G}=\{\mathcal{G}^e\}_e$ collected from multiple environments $\mathcal{E}_{\text{all}}$, each representing a different domain or data-generating condition. Each environment $\mathcal{E}$ corresponds to a distribution $\mathcal{P}_{\mathcal{E}}$ from which graph-label pairs $(G, Y)$ are sampled. For each dataset $\mathcal{G}^e = \{(G_i^e, Y_i^e)\}_{i=1}^{N^e}$, $(G_i^e, Y_i^e)$ is drawn from the distribution of environment $e$. In the graph OOD setting, training and testing environments—denoted as $\mathcal{E}_{tr}$ and $\mathcal{E}_{te}$—are assumed to differ, creating a distribution shift. Moreover, the environment labels are often unobserved in practice due to the high cost of manual annotation, making environment-agnostic or environment-inference-based approaches more desirable.

\subsection{Graph-level Invariant Learning}
Graph Invariant Learning (GIL)~\cite{li2022learning} aims to learn an environment-agnostic function $f: \bigcup \mathcal{G} \rightarrow \mathcal{Y}$ that predicts labels based on global graph-level representations, invariant across environments. The overall goal is:
\begin{equation}\label{eq:problem_formulate}
f^* = \arg \min_{f} \max_{e\in\mathcal{E}_{\text{all}}} \mathcal{R}(f \vert e),
\end{equation}
where $\mathcal{R}(f \vert e) = \mathbb{E}_{(G, Y)\in\mathcal{G}^e} [\ell(f(G), Y)]$ is the expected risk in environment $e$ and $\ell: \mathcal{Y} \times \mathcal{Y} \rightarrow \mathbb{R}$ is the loss function.

Several modern approaches have pursued this direction. \cite{GraphKDD22} proposes a rationale-driven augmentation framework that separates spurious and invariant subgraphs to support more robust training. \cite{chen2023rethinking} introduces an environment-free invariant learning paradigm based on mutual information maximization, avoiding the need for manual environment partitioning. SGooD~\cite{ding2024sgood} incorporates substructure priors for graph-level OOD detection, while DIVE~\cite{sun2024dive} explores subgraph disagreement signals across environments to enhance generalization. In~\cite{jia2024graph}, the authors combine subgraph co-mixup with invariant risk minimization to improve graph-level robustness. The recent effort~\cite{liu2025subgraph} further explores subgraph aggregation as a mechanism to align invariant features across domains.

While these works contribute to learning stable graph-level representations, they still rely on discrete masking, predefined substructures, or hand-crafted augmentations which can be hard and even ad-hoc. Technically, they can hardly provide mechanisms to control \emph{separability}, implement \emph{soft attention}, or ensure full \emph{differentiability} in the subgraph extraction process as fulfilled by our work. It limits flexibility and the process is component-wise.  In contrast, we try to provide a unified and principled end-to-end approach without resorting to specific rules.

\subsection{Node-level Invariant Learning}
Node-level invariant learning, although less studied, is crucial for large-scale graphs and real-world applications e.g. citation networks, transaction prediction, and fraud detection. The main challenge lies in capturing node-centered patterns that remain stable across graph environments.

Recent efforts have started addressing this gap. The work~\cite{wang2025subgraph} propose a subgraph-invariant learning framework specifically designed for node classification tasks, leveraging local neighborhood extraction and global alignment. In~\cite{chen2023does}, it is questioned whether environment augmentation-based training indeed leads to invariance at the node level, motivating more structured invariance modeling. Our work follows this direction by introducing a differentiable, separability-controllable mechanism that softly selects node-relevant subgraphs and generalizes to unseen environments.
Compared to approaches that rely on either global pooling or hand-crafted masking, our method provides fine-grained, soft, and trainable subgraph extraction tailored for node-level generalization.

\subsection{Subgraph-based Invariant Learning}
Since the problem in Eq.~\ref{eq:problem_formulate} is difficult as the environment variable $e$ is supposed unobserved, a line of research has emerged that focuses on subgraphs, with the goal of identifying an invariant subgraph of the input, bearing a stable relationship to the label; and filtering out the other part of the input, which is environment-relevant or spurious. 
% We discuss the subgraph-based methods in two categories: i) extracting subgraphs, ii) working with given subgraphs.

% \subsubsection{Learning with Subgraph Extraction}
There are existing approaches that aim to explicitly extract invariant subgraphs (with various definitions), guided by the information bottleneck (IB) principle \cite{yu2020graph,miao2022interpretable} or top-$k$ selection \cite{wu2022discovering,chen2022learning}. In particular, \cite{miao2022interpretable} introduces Graph Stochastic Attention (GSAT), an attention mechanism that constructs inherently interpretable and generalizable GNNs. This attention is formulated as an information bottleneck by introducing stochasticity into the attention mechanism, which constrains the information flow from the input graph to the prediction. By penalizing the amount of information from the input data, GSAT is expected to be more generalizable. \cite{chen2022learning} proposes a Causality Inspired Invariant Graph LeArning (CIGA) framework to capture the invariance of graphs under various distribution shifts. Specifically, they characterize potential distribution shifts on graphs with causal models, which focus only on subgraphs containing the most information regarding the causes of labels. Overall, these approaches provide exciting opportunities for achieving interpretability and generalizability in GNNs without requiring expensive domain labels. The work~\cite{li2022learning} also proposes an invariant subgraph learning scheme based on complementarity and clustering. These works do not include any discussion or mechanism for enforcing separability or softness, nor does it offer a differentiable formulation for subgraph selection. Another related work refers to~\cite{yang2022learning} which makes an important observation to the relation between the properties and substructures of the molecules. Specifically, the robustness for graph-level prediction is expected to be achieved by leveraging the substructure information instead of the whole molecule.
% \subsubsection{Learning with given Subgraphs}
These works also do not involve subgraph extraction (instead the subgraph is supposed to be given). To automatically extract the subgraphs, the work~\cite{DebiasingNIPS22} devises an edge mask-based mechanism to split causal sugraphs from the spurious subgraphs. In their approach, the edge masking is discrete, non-differentiable. In~\cite{PR25}, the authors propose the so-called joint subgraph independence to jointly eliminates the spurious correlations between subgraphs and correlations between subgraphs and external factors.

More recently, \cite{chenICML24interpretable} introduces the multilinear extension of the subgraph distribution and formulates the interpretable subgraph learning framework called subgraph multilinear extension (SubMT). Then a new architecture called Graph Multilinear net is devised to approximating the inductive bias i.e., SubMT with the hope of obtaining better interpretability. The experiments are mainly performed on graph classification yet the approach still faces scalability issue for node level tasks as emphasized by the authors in the paper. Another robust and general inductive bias is devised in \cite{yao2024empowering} based on the observation that the infomax principle encourages learning spurious features. Then a new objective is designed to elicit invariant features by disentangling them from the spurious features learned via infomax. The approach is designed for graph level classification. In~\cite{TianjunICLR25Spuriosity} for graph classification, it first learns the spurious features and then removes them from the feature pool learned by the classic Empirical Risk Minimization principle.

% Therefore, we introduce the concept of invariant subgraph $G_S$, and focus on inferring and filtering out the environment-relevant part $\overline{G_S}$ and keeping the label-relevant (invariant) part $G_S$. 
% We use the mathematic tool of mutual information to formulate this problem as described in Sec.~\ref{sec:learning_obj}.

% The concept of invariant learning involves utilizing the invariant relationships between features and labels across various distributions while disregarding any spurious correlations that may arise \cite{li2022learning}. 
% Through this approach, it is possible to attain a high level of out-of-distribution (OOD) generalization in the presence of distribution shifts. 

% Despite the invariance principle's potential for achieving out-of-distribution (OOD) generalization, there has been a lack of comprehensive exploration of its application to graph data. 
% However, this area necessitates further research to fully understand graph invariant learning for graph data OOD generalization. 

% Existing methods for domain adaptation in graphs typically require expensive domain labels, which limits their practical applications. 

%Handling the constraint violation is the core of the cardinality-constraint CO problem as a tighter constraint violation leads to better performance
%%\vspace{-3pt}
%with wide applications \cite{chen2021network,chang2000heuristics}
\subsection{Cardinality-based Combinatorial Optimization}\label{sec:card}
%%\vspace{-3pt}
The presented technique in this paper also bears connection to those used in recent learning-based combinatorial optimization (CO)~\cite{ausiello2012complexity}. Specifically, cardinality-constraint optimization is a permutation-based CO problem, whose solution includes at most $k$ non-zero entries, i.e., the cardinality constraint $\Vert\mathbf{x}\Vert _0 \leq k$. Choosing the top-$k$ most influential edges for the invariant subgraph, which is a constraint-critical scenario, could also be regarded as a cardinality-constraint problem~\cite{wang2023towards}. Erdos Goes Neural~\cite{karalias2020erdos} places a penalty term for a constraint violation in the loss, but the constraint violation is unbounded. \cite{xie2020differentiable} develops a soft algorithm by recasting the top-$k$ selection as an optimal transport problem~\cite{ot_tpami24} with the Sinkhorn algorithm~\cite{Sinkhorn1964ARB}. Although the upper limit of constraint violation is provided, in the worst situation, the bound might diverge. \cite{wang2023towards} further addresses the issue and proposes a method with a tighter upper bound by introducing the Gumbel trick, making the constraint violation arbitrarily controlled. %These studies could provide a theoretical basis to formulate the problem of GIL from the perspective of combinatorial optimization, which also inspire our approach.

\section{Approach}\label{sec:method} 
% In this section, we will first introduce the optimization, or learning objective of our Graph Sinkhorn Attention (GSINA) for Graph Invariant Learning (GIL) at a high level in Sec.~\ref{sec:learning_obj}, then the implementation details of GSINA: the utilization of the Sinkhorn algorithm to obtain the sparse, soft, and differentiable invariant subgraph as a kind of graph attention mechanism, and the general representation learning framework for GIL of multiple level tasks in Sec.~\ref{sec:GSA}. 

\begin{figure}[tb!]%{R}{0.4\textwidth}
% %%\vspace{-20pt}
    \centering
    \includegraphics[width=0.96\linewidth]{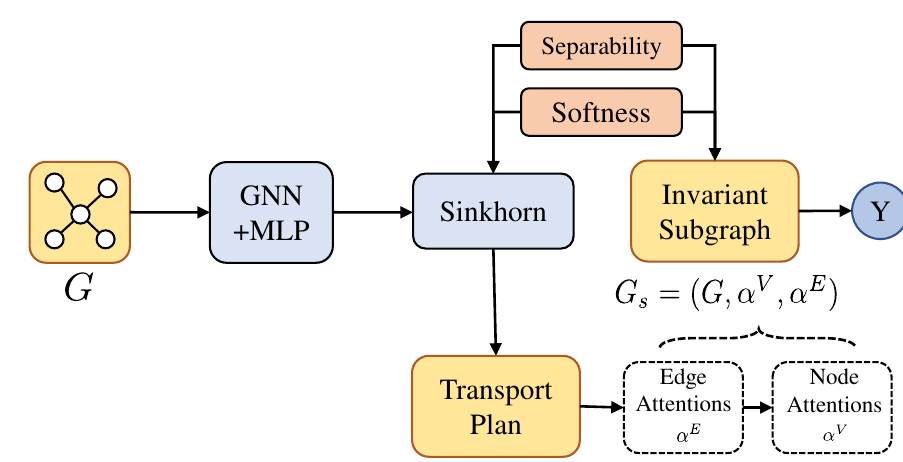}
    % %%\vspace{-5pt}
    \caption{Overview of our approach. A separable invariant subgraph is extracted from a given input graph using Sinkhorn-based optimal transport with controllable separability and softness. Edge and node attention scores are computed to form the subgraph, which are then fed into a predictor to generate the output label $Y$  for nodes and graphs.
    }
    \label{fig:pipeline}
    %%\vspace{-10pt}
\end{figure}

% \hzp{
We first introduce the learning objective of our Graph Sinkhorn Attention (GSINA) for Graph Invariant Learning (GIL) at a high level in Sec.~\ref{sec:learning_obj}. 
Then in Sec.~\ref{sec:GSA} we present the implementation details of GSINA: the utilization of the Sinkhorn algorithm to obtain the separable, soft, and differentiable invariant subgraph as a kind of graph attention mechanism, and the general representation learning framework for GIL of multiple-level tasks. The theoretical study on the convergence of our algorithm is presented in Sec.~\ref{sec:converge}. An overview of our approach is shown in Fig.~\ref{fig:pipeline}.
% }

% 
%%\vspace{-3pt}
\subsection{Graph Sinkhorn Attention: Optimization}\label{sec:learning_obj}
%%\vspace{-3pt}
Aiming at finding the invariant subgraph $G_S$ with a stable relationship to the label $Y$, we formulate it as a mutual information $I(G_S ; Y)$ maximization problem like the practices in~\cite{yu2020graph,miao2022interpretable,chen2022learning}. On the other hand, constraints should be applied on the invariant subgraph $G_S$ to ensure its informative conciseness (i.e., the information of the variant or redundant part of the input graph $G$ should be damped), the constraints act as regularizations and improve the generalization of graph learning tasks. As discussed in Sec.~\ref{sec:intro}, the information bottleneck (IB) based methods~\cite{yu2020graph,miao2022interpretable} might hardly guarantee the subgraph conciseness (lack of separability), and the partially differentiable top-$k$ selection based methods~\cite{wu2022discovering,chen2022learning} generate hard subgraphs and shrink the subgraph solution space. Differently, we leverage a subgraph extractor $g_{\phi}(G, r, \tau)$ generating softly cardinality-constrained subgraph $G_S$, and the subgraph ratio as defined by $r$ in this paper, is the cardinality constraint controlling the separability of $G_S$. While $\tau$ is a temperature hyperparameter controlling the softness of $G_S$, which will be detailedly discussed in Sec.~\ref{sec:GSA}. 

In fact, our subgraph extractor $g_{\phi}(G, r, \tau)$ acts as a separability and softness regularization of $G_S$, and the mutual information maximization problem can be formulated as:
\begin{equation} \label{eq:MI}
    \max_{\phi} I\left(G_S ; Y\right),\, \text{s.t.}\,\, G_S \sim g_{\phi}(G, r, \tau). 
\end{equation}

As a direct estimation of mutual information  $I(G_S ; Y)$ is intractable, we derive its lower bound with the help of a variational approximation distribution $P_{\theta}(Y|G_S)$ parameterized by $\theta$, which also acts as the predictor of label $Y$ given the invariant subgraph $G_S$:
\begin{equation}\label{eq:lb}
\begin{aligned} 
&I\left(G_S ; Y\right) = \iint_{G_S, Y} P(G_S, Y) \log\frac{P(G_S, Y)}{P(G_S)P(Y)} \d G_S \d Y \\ 
% &= \iint_{G_S, Y} P(G_S, Y) \log P(Y | G_S) \d G_S \d Y + H(Y) \\ 
&= \int_{G_S} P(G_S)\int_{Y} P(Y | G_S) \log P_{\theta}(Y | G_S) \d Y  \\&\quad +KL[P(Y | G_S) \Vert P_{\theta}(Y | G_S)] \d G_S + H(Y) \\ 
&\geq \int_{G_S} P(G_S)\int_{Y} P(Y | G_S) \log P_{\theta}(Y | G_S) \d Y \d G_S + H(Y) \\
% &=\iint_{G_S, Y} P(G_S, Y) \log P_{\theta}(Y | G_S) \d Y  \d G_S + H(Y) \\
&= \mathbb{E}_{G_S, Y} \left[ \log {P_{\theta}(Y|G_S)} \right] + H(Y),
\end{aligned}
\end{equation}
where $H(Y)=-\int_{Y} P(Y) \log P(Y) \d Y$ is a constant entropy of the label distribution $P(Y)$ and can be omitted in the optimization. Then the problem in Eq.~\ref{eq:MI}  can be optimized by maximizing the lower bound item in Eq.~\ref{eq:lb} and the final learning objective of GSINA is:
\begin{align}\label{eq:lo} %\nonumber  
    &\max_{\theta,\phi}\,\mathbb{E}_{G_S, Y}\left[ \log {P_{\theta}(Y|G_S)} \right],\,  \text{s.t.}\,\,  G_S \sim g_{\phi}(G, r, \tau).
\end{align}

It leads to a differentiable pipeline: first extracting the invariant subgraph $G_S$ from the input graph $G$, then making prediction $Y$ based on $G_S$. Though sharing some similarities, GSINA is unlike the IB-based methods~\cite{yu2020graph,miao2022interpretable}: the subgraph information is constrained by our subgraph extractor $g_{\phi}(G, r, \tau)$, able to explicitly control the separability ratio $r$ and softness $\tau$ of the invariant subgraph.

% (as shown in Fig.~\ref{fig:pipeline})

%%\vspace{-3pt}
\subsection{Graph Sinkhorn Attention: Implementation}\label{sec:GSA}
%%\vspace{-3pt}
To extract a separable, soft, and differentiable invariant subgraph $G_S$ from the input graph $G$, we leverage a softly cardinality-constrained subgraph extractor $g_{\phi}(G, r, \tau)$ based on the Sinkhorn algorithm~\cite{Sinkhorn1964ARB}.

Recall that the goal of graph attention mechanisms~\cite{GAT,miao2022interpretable} is to assign attention coefficients to different parts of the input graph structure, evaluating their respective importance for the prediction of the target label $Y$. Beyond the graph attention mechanisms designed in~\cite{GAT,miao2022interpretable}, we take the separability and softness of the attention distribution into consideration by applying differentiable top-$k$~\cite{xie2020differentiable,wang2023towards} to evaluate the importance of edges.  Recall the cardinality-constrained combinatorial optimization described in Sec.~\ref{sec:card}, it has a corresponding popular OT-theoretic solution of the Sinkhorn algorithm~\cite{Sinkhorn1964ARB}, and the Gumbel re-parameterization trick could be readily adopted to enhance the performance according to the practice in~\cite{wang2023towards}.

\begin{figure}[tb!]%{R}{0.4\textwidth}
% %%\vspace{-20pt}
    \centering
    \includegraphics[width=0.44\textwidth]{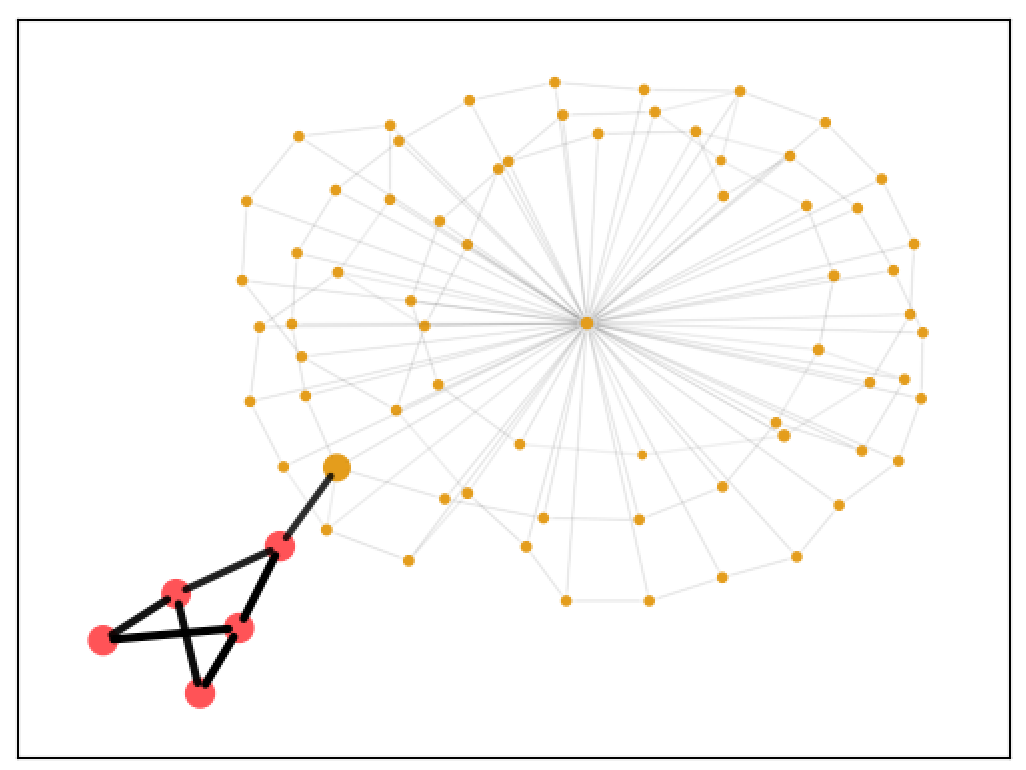}
    % %%\vspace{-5pt}
    \caption{Example of GSINA invariant subgraph $G_S$ from SPMotif dataset. The ground truth (nodes) of the invariant subgraph is colored red, and the other part is yellow. The edge widths and node sizes are given by GSINA original outputs (we do not apply sharpening tricks for visualizations like GSAT~\cite{miao2022interpretable} and CIGA~\cite{chen2022invariance}). It is shown that our GSINA assigns sparse and soft attention $ \{\alpha^V, \alpha^E \}$ to the nodes and edges of the input graph $G$.   More visualization results can be found in \url{https://github.com/dingfangyu/GSINA/vis}. }
    \label{fig:gsa}
    %%\vspace{-10pt}
\end{figure}

Our Graph Sinkhorn Attention leverages a cardinality-constrained subgraph extractor to generate sparse and soft attention coefficients for edges and nodes of the input graph $G$. For edge attention, it softly highlights the top-$r$ ratio most influential edges and `filters out' other edges by assigning sparse edge attention to the input graph $G$ (see Fig.~\ref{fig:demo} and Fig.~\ref{fig:gsa}), and provides soft attention distribution. Based on GSINA edge attention, sparse and soft node attention could be designed based on graph neighborhood aggregation to evaluate the node importance.

We will start by describing the implementation of edge and node attention in GSINA, and then the general framework for multiple-level GIL tasks via GSINA. 

\textbf{Edge Attention.} 
 As an initial step, a composition of $\text{GNN}_{\phi}$ and $\text{MLP}_{\phi}$ is leveraged to obtain learnable node features $\left\{\mathbf{n}_i |i \in \mathcal{V}  \right\}$ and edge scores $s = \{s_e | e \in \mathcal{E}\}$ of the input graph $G = (\mathcal{V}, \mathcal{E})$:
 \begin{equation}
\begin{aligned}\label{eq:gnn1}
    \left\{\mathbf{n}_i \right\} &= \text{GNN}_{\phi} (G), \,\, i \in \mathcal{V},  \\
    s_e\,\,\,  &= \text{MLP}_{\phi} (\mathbf{n}_i, \mathbf{n}_j), \,\, e = (i, j).
\end{aligned}
 \end{equation}

\begin{algorithm}[tb!]
\caption{The training procedure for GSINA.}\label{alg:train}
\textbf{Hyperparameters}: the number of training epoch $E$; the number of batch size $B$; the separability $r$; the softness $\tau$; the Sinkhorn iterations $n$.

\textbf{Input}: training dataset $\mathcal{G} = \{G_i, Y_i\}_{i}^N$ with paired labels.

\textbf{Output}: the trained parameters $\theta$ and $\phi$ in Sec.~\ref{sec:learning_obj}.

\begin{algorithmic}[1]
\STATE Initialize parameters $\theta$ and $\phi$;\\
{// \texttt{For each epoch}}
\FOR{$i = 1,\dots, E$}
    \STATE Sample data batches $\mathcal{B} = \{\mathcal{G}_1, \mathcal{G}_2, \dots , \mathcal{G}_b\}$ from $\mathcal{G}$ with batch size $B$;\\
    {// \texttt{For each batch}}
    \FOR{$j = 1 ,\dots, b$}
        \STATE $G = \{G_m | (G_m,Y_m)\in \mathcal{G}_j\}$; 
        \STATE $Y = \{Y_m | (G_m,Y_m)\in \mathcal{G}_j\}$;\\
        %\STATE Unfold $\mathcal{G}_j = (G, Y)$;
        {// \texttt{Forward: get $G_S = \{G,\alpha^V, \alpha^E \} \sim g_{\phi}(G, r, \tau)$ by Eq.~\ref{eq:Gs}}} 
        \STATE Get edge scores $s = \{s_e | e \in \mathcal{E}\}$  with $\text{GNN}_{\phi}$ and $\text{MLP}_{\phi}$  by Eq.~\ref{eq:gnn1};
        \STATE Prepare $\mathbf{D}, \mathbf{R}, \mathbf{C}$ of the OT problem by Eq.~\ref{eq:ot};
        \STATE Initialize $\mathbf{T}_0$ by Eq.~\ref{eq:sinkhorn0};
        \FOR{$k = 1,\dots,n$}
            \STATE Update $\mathbf{T}_k$ by Eq.~\ref{eq:sinkhorn1} and~\ref{eq:sinkhorn2};
        \ENDFOR
        \STATE Get edge/node attention $\{\alpha^V, \alpha^E\}$ by Eq.~\ref{eq:edge_att} and~\ref{eq:node_att}; \\      % \STATE Get predicted label $\hat{Y} = P_\theta(Y|G_S)$ according to Eq.~\ref{eq:msg_pass}; 
        {//\texttt{Forward: get the objective $\log {P_{\theta}(Y|G_S)}$ in Eq.~\ref{eq:lo} by Eq.~\ref{eq:msg_pass} and Eq.~\ref{eq:msg_pass2}}}
        \IF{Graph-level GIL task}
            \STATE Get  $h_G$  by Eq.~\ref{eq:msg_pass2}; 
            \STATE Model the distribution $P_{\theta}(Y|G_S)$ with $h_G$;
        \ELSIF{Node-level GIL task}
            \STATE Get  $\{h^{L}_v | v\in\mathcal{V}\}$  by Eq.~\ref{eq:msg_pass}; 
            \STATE Model distribution $P_{\theta}(Y|G_S)$ by $\{h^{L}_v | v\in\mathcal{V}\}$;
        \ENDIF
        \STATE Get the learning objective $\log {P_{\theta}(Y|G_S)}$ in Eq.~\ref{eq:lo};
        \\{// \texttt{Backward}}
        \STATE Optimize parameters $\theta$ and $\phi$ via gradient descent;
    \ENDFOR
\ENDFOR
\STATE Output the parameters $\theta$ and $\phi$;
\end{algorithmic}
\end{algorithm}

Softly selecting the top-$r$ scored edges as the invariant subgraph could be interpreted as a relaxed OT problem, whose setting is to move $r N_e$ items to the destination of the invariant part, and the other $(1-r) N_e$ elements to the variant part, where $N_e$ is the number of the edges. 
During training, the Gumbel re-parameterization trick
$\Tilde{s_e} = s_e  - \sigma \log(-\log u_e), u_e \sim U(0,1)$~\cite{jang2016categorical,wang2023towards}, where $\sigma$ is the factor of Gumbel noise and $\sigma=0$ in validation and testing phases, could be adopted to enlarge the sampling space and to improve the generalization performances, i.e. the Gumbel trick allows less important edges to participate in training (without being poorly trained due to low attention, resulting in underfitting), and remains the sampling accuracy. Defining $\mathbf{D}$ as the distance matrix of the OT problem, $\mathbf{R}$ and $\mathbf{C}$ as the marginal distributions, and $\mathbf{T}$ as the transportation plan moving $r N_e$ items to $\text{max}(s)$ (invariant) and $(1-r) N_e$ items to $\text{min}(s)$ (variant), the OT problem for GSINA edge attention can be formulated as follows:
\begin{align}
%\begin{gathered}
\label{eq:ot}
    \mathbf{D} = &\begin{bmatrix}
        \Tilde{s_1} - \min(s), &\Tilde{s_2} - \min(s), &\dots,  &\Tilde{s}_{N_e} - \min(s)\\
        \max(s) - \Tilde{s_1}, &\max(s) - \Tilde{s_2}, &\dots,  &\max(s) - \Tilde{s}_{N_e}
        \end{bmatrix},\\
   \mathbf{R} =& [(1 -r) N_e ,\,  r N_e]^\top , \,\mathbf{C} = [1,\, 1,\, \dots,\, 1]^\top \in \mathbb{R}^{N_e \times 1}, \\\label{eq:eot}
   & \min_{\mathbf{T}} \text{tr}(\mathbf{T}^\top \mathbf{D})  - \tau H(\mathbf{T}), \\ 
    &\text{s.t.}\quad 
    \mathbf{T}\in [0, 1]^{2 \times N_e } ,\,\, \mathbf{T} \mathbf{1} = \mathbf{R},\,\, \mathbf{T}^\top \mathbf{1}= \mathbf{C}, 
%\end{gathered}
\end{align}
where $\tau H(\mathbf{T})$ is the entropic regularizer~\cite{NIPS2013_af21d0c9} of the OT problem, and $\tau$ is the temperature hyperparameter controlling the softness of the transportation plan $\mathbf{T}$. The result of $\mathbf{T}$ could be iteratively calculated by the Sinkhorn scheme:
% \begin{equation}
\begin{align}%\label{eq:sinkhorn}
    %\begin{align}
    \mathbf{T}_0 = &\exp\left(-\frac{\mathbf{D}}{\tau}\right),\label{eq:sinkhorn0}\\
    \mathbf{T}_k = &\operatorname{diag}\left(\mathbf{T}_{k-1}\mathbf{1} \oslash \mathbf{R}\right)^{-1}\mathbf{T}_{k-1},\label{eq:sinkhorn1}\\ \mathbf{T}_k = &\mathbf{T}_{k-1}\operatorname{diag}(\mathbf{T}_{k-1}^\top \mathbf{1} \oslash \mathbf{C}) ^{-1},\label{eq:sinkhorn2}
    %\end{aligned}
\end{align}
% \end{equation}
where $\mathbf{T}_0$ in Eq.~\ref{eq:sinkhorn0} is the initialization. The equations of $\mathbf{T}_k$ in Eq.~\ref{eq:sinkhorn1} and Eq.~\ref{eq:sinkhorn2} are alternative iterations of row- and column-wise normalizations to satisfy the two constraints $\mathbf{T} \mathbf{1} = \mathbf{R}$ and $\mathbf{T}^\top \mathbf{1}= \mathbf{C}$, $\oslash$ is element-wise division.
 
Eventually, the edge attention $\alpha^E = \{\alpha^E_e | e\in \mathcal{E}\}$ could be obtained from the procedure above:
\begin{equation}\label{eq:edge_att}
    \left[\alpha^E_1,\,\alpha^E_2,\,\dots,\alpha^E_{N_e}\right] = \mathbf{T}[1, :].
\end{equation}

% \subsectionGSINA
\textbf{Node Attention.}
Given the sparse, soft, and differentiable edge attention $\alpha^E$, a natural consideration is to evaluate the importance of nodes. Hence, the node attention $\alpha^V =\{\alpha^V_i | i\in \mathcal{V}\}$ in our GSINA is proposed, which could be obtained by an aggregation of the edge attention in the neighborhood of each node $i$:
\begin{equation}\label{eq:node_att}
    \alpha^V_i = \text{AGG}(\{\alpha^E_{e} \vert e=(i, j) \land e\in \mathcal{E}\}).
\end{equation}

For the `hard' top-$k$  based GIL methods~\cite{wu2022discovering, chen2022learning}, only the selected part is kept to be the invariant subgraph $G_S$ and the other part $\overline{G_S}$ is usually discarded. In other words, the node importance is 1 for nodes in $G_S$ and 0 for nodes in $\overline{G_S}$. In contrast, our node attention is a soft and fully differentiable version to mimic their invariant subgraph extractions.

\textbf{General Graph Invariant Learning Framework.} Our invariant subgraph extraction results in a
 graph weighted by the Graph Sinkhorn (Edge and Node) Attention, with the properties of separability, softness, and differentiability, the mathematical definition of the invariant subgraph $G_S$ of our GSINA is in the manner of graph attention:
\begin{equation}\label{eq:Gs}
    G_S = \{G,\alpha^V, \alpha^E \} \sim g_{\phi}(G, r, \tau).
\end{equation}

\begin{table*}[tb!]
    \centering
    %%\vspace{-10pt}
    \caption{{Dataset statistics of the GSAT benchmark used in Table~\ref{tab:gsat}, following the datasets used in DIR~\cite{wu2022discovering}.}}
    \label{tab:gsat-data-1}
    % \resizebox{1.0\textwidth}{!}
    {
        \begin{tabular}{rccccccccccccc}
    \toprule
     &  \multicolumn{3}{c}{Spurious-Motif} &\multicolumn{3}{c}{MNIST-75sp (reduced)}& \multicolumn{3}{c}{Graph-SST2}& \multicolumn{3}{c}{OGBG-Molhiv}\\ 
     & Train & Val & Test & Train & Val & Test & Train & Val & Test& Train & Val & Test\\ \midrule
    Classes\# & &3 & & &10&   &  &2& &&2&\\
    Graphs\#  & 9,000& 3,000& 6,000& 20,000 & 5,000 & 10,000 & 28,327 & 3,147&  12,305 &32,901&4,113&4,113\\
    Avg. N\#  & 25.4 & 26.1 & 88.7& 66.8  & 67.3 & 67.0  & 17.7  & 17.3  & 3.45&25.3& 27.79 & 25.3\\
    Avg. E\#  & 35.4 & 36.2 & 131.1& 539.3 & 545.9 & 540.4 & 33.3 & 33.5 & 4.89&54.1& 61.1 & 55.6\\
    Metrics & &ACC & & &ACC&   &  &ACC& &&ROC-AUC&\\
    \bottomrule
    \end{tabular}}
\end{table*}

\begin{table*}[tb!]
%%\vspace{-5pt}
\caption{Comparison with the IB based GIL, graph-level OOD generalization performances on GSAT~\cite{miao2022interpretable} benchmark.}
%%\vspace{-10pt}
\begin{center}
% \begin{small}
% \resizebox{\columnwidth}{!}{
% \begin{sc}
\begin{tabular}{lcccccc}
\toprule
  & \multirow{2}{*}{MolHiv (AUC)} & \multirow{2}{*}{Graph-SST2} & \multirow{2}{*}{MNIST-75sp} & \multicolumn{3}{c}{Spurious-motif}                                        \\
  &                    &                    &                    & $b=0.5$                  & $b=0.7$                  & $b=0.9$                  \\
\midrule
GIB~\cite{yu2020graph} & ${76.43}{\scriptstyle \pm2.65}$ & $82.99{\scriptstyle \pm0.67}$ & $93.10{\scriptstyle \pm1.32}$ & $54.36{\scriptstyle \pm7.09}$ & $48.51{\scriptstyle \pm5.76}$ & $46.19{\scriptstyle \pm5.63}$ \\
DIR~\cite{wu2022discovering} & $76.34{\scriptstyle \pm1.01}$ & $82.32{\scriptstyle \pm0.85}$ & $88.51{\scriptstyle \pm2.57}$ & $45.49{\scriptstyle \pm3.81}$ & $41.13{\scriptstyle \pm2.62}$ & $37.61{\scriptstyle \pm2.02}$ \\
% CIGA &$73.00{\scriptstyle \pm1.27}$&$82.00{\scriptstyle \pm0.79}$&$-$&$67.30{\scriptstyle \pm7.90}$&$57.47{\scriptstyle \pm15.2}$& $48.76{\scriptstyle \pm15.5}$ \\
\midrule
GIN~\cite{xu2018powerful} & $76.69{\scriptstyle \pm1.25}$ & $82.73{\scriptstyle \pm0.77}$ & $95.74{\scriptstyle \pm0.36}$ & $39.87{\scriptstyle \pm1.30}$ & $39.04{\scriptstyle \pm1.62}$ & $38.57{\scriptstyle \pm2.31}$ \\

%\midrule
GIN+GSAT~\cite{miao2022interpretable} & $76.47{\scriptstyle \pm1.53}$ & $82.95{\scriptstyle \pm0.58}$ & $96.24{\scriptstyle \pm0.17}$ & $52.74{\scriptstyle \pm4.08}$ & $49.12{\scriptstyle \pm3.29}$ & $44.22{\scriptstyle \pm5.57 }$\\
 GIN+GSINA (ours) & 
$\mathbf{77.99}{\scriptstyle \pm0.97}$ & % r=0.9
$\mathbf{83.66}{\scriptstyle \pm0.37}$ & % r = 0.2
$\mathbf{96.73}{\scriptstyle \pm0.16}$ & % r = 0.5
$\mathbf{55.16}{\scriptstyle \pm5.69}$ & % r=0.6
$\mathbf{56.83}{\scriptstyle \pm6.32}$ &  %r=0.6
$\mathbf{49.86}{\scriptstyle \pm6.10}$\\ % r=0.4
%  w/o Gumbel & 
% $-$ & 
% $83.45{\scriptstyle \pm0.42}$ & 
% $96.87{\scriptstyle \pm0.22}$ & $48.27{\scriptstyle \pm4.80}$ & $45.25{\scriptstyle \pm7.15}$ & $50.28{\scriptstyle \pm2.83}$  \\ 

%  w/o NodeAttn & 
% $-$ & 
% $83.41{\scriptstyle \pm1.09}$ & $96.47{\scriptstyle \pm0.46}$ & $47.34{\scriptstyle \pm7.99}$ & $54.63{\scriptstyle \pm6.99}$ & $48.41{\scriptstyle \pm1.16}$  \\ 
%  w/o G\&N & $-$ & 
% $84.07{\scriptstyle \pm0.38}$ & $96.87{\scriptstyle \pm0.28}$ & $46.28{\scriptstyle \pm5.67}$ & $45.40{\scriptstyle \pm3.22}$ & $44.44{\scriptstyle \pm6.56}$  \\ 
\midrule
PNA~\cite{corso2020principal} & $78.91{\scriptstyle \pm1.04}$ & $79.87{\scriptstyle \pm1.02}$ & $87.20{\scriptstyle \pm5.61}$ & $68.15{\scriptstyle \pm2.39}$ & $66.35{\scriptstyle \pm3.34}$ & $61.40{\scriptstyle \pm3.56} $ \\
PNA+GSAT~\cite{miao2022interpretable} & $80.24{\scriptstyle \pm0.73}$ & $80.92{\scriptstyle \pm0.66}$ & $93.96{\scriptstyle \pm0.92}$ & $68.74{\scriptstyle \pm2.24}$ & $64.38{\scriptstyle \pm3.20}$ & $57.01{\scriptstyle \pm2.95} $\\
 PNA+GSINA (ours) & 
$\mathbf{80.55}{\scriptstyle \pm0.97}$ & % r=0.7 
$\mathbf{82.18}{\scriptstyle \pm1.01}$ & % r = 0.8
$\mathbf{95.48}{\scriptstyle \pm0.37}$ & % r = 0.6
$\mathbf{76.39}{\scriptstyle \pm1.85}$ & % r=0.1
$\mathbf{73.96}{\scriptstyle \pm2.87}$ &  %r=0.3
$\mathbf{62.51}{\scriptstyle \pm5.86}$\\ % r=0.5
% w/o Gumbel & 
% $-$ & 
% $82.12{\scriptstyle \pm0.98}$ & $94.59{\scriptstyle \pm0.98}
% $ & $69.95{\scriptstyle \pm2.76}$ & $69.67{\scriptstyle \pm3.44}$ & $62.14{\scriptstyle \pm4.64}$  \\ 

%  w/o NodeAttn & 
% $-$ & 
% $81.42{\scriptstyle \pm0.77}$ & $95.51{\scriptstyle \pm0.18}$ & $71.60{\scriptstyle \pm1.89}$ & $58.70{\scriptstyle \pm3.82}$ & $58.20{\scriptstyle \pm2.70}$  \\ 

%  w/o G\&N & 
% $-$ & 
% $82.14{\scriptstyle \pm1.05}$ & $93.98{\scriptstyle \pm1.38}$ & $71.75{\scriptstyle \pm2.42}$ & $67.50{\scriptstyle \pm4.51}$ & $61.34{\scriptstyle \pm1.72}$  \\ 
\bottomrule
\label{tab:gsat}
\end{tabular}

% \end{sc}
% }
% \end{small}
\end{center}
     %%\vspace{-20pt}
\end{table*}

\newcommand{\mr}[2]{\multirow{#1}{*}{#2}}
\begin{table}[tb!]
  \centering
  %%\vspace{-10pt}
  \caption{{ Statistics of ogbg-mol* of GSAT benchmark in Table~\ref{tab:gsat_ogbg}.} We use the scaffold split with split ratio 8:1:1. All the tasks are binary classification with ROC-AUC as the metric.}
  %%\vspace{-5pt}
  \label{tab:gsat-data-2}
  \renewcommand{\arraystretch}{1}
  %\setlength{\tabcolsep}{3pt}
% \resizebox{\linewidth}{!}{
  \begin{tabular}{lrrrr}
    \toprule
        {\textbf{dataset}} & {\textbf{\#Graphs}} & \textbf{\#Nodes}     & \textbf{\#Edges} &   {\textbf{\#Tasks}}  \\
    \midrule
       \texttt{bace} & 1,513 & 34.1 & 36.9 & 1   \\
       \texttt{bbbp} & 2,039 & 24.1 & 26.0 & 1   \\
       \texttt{clintox} & 1,477 & 26.2 & 27.9 & 2   \\
       \texttt{tox21} & 7,831 & 18.6 & 19.3 & 12   \\
       \texttt{sider} & 1,427 & 33.6 & 35.4 & 27   \\
      % & \texttt{esol} & 1,128 & 13.3 & 13.7 & 1 & Regression & RMSE  \\
      % & \texttt{freesolv} & 642 & 8.7 & 8.4 & 1 & Regression & RMSE  \\
      % & \texttt{lipo} & 4,200 & 27.0 & 29.5 & 1 & Regression & RMSE  \\
    \bottomrule
  \end{tabular}
 %}
\end{table}
% \begin{wraptable}{r}{0.65\textwidth}
\begin{table}[tb!]
%%\vspace{-5pt}
\caption{Comparison of IB-based GIL, graph-level OOD generalization on other OGBG-Mol datasets in GSAT~\cite{miao2022interpretable}.}%%\vspace{-10pt}
\begin{center}
\resizebox{\columnwidth}{!}{%
% \begin{sc}
\begin{tabular}{lccccc}
\toprule
  & molbace           & molbbbp           & molclintox        & moltox21          & molsider          \\
\midrule
PNA~\cite{corso2020principal} & $73.52{\scriptstyle \pm3.02}$ & $67.21{\scriptstyle \pm1.34}$ & $86.72{\scriptstyle \pm2.33}$ & $75.08{\scriptstyle \pm0.64}$ & $56.51{\scriptstyle \pm1.90}$ \\
GSAT~\cite{miao2022interpretable}            & ${77.41}{\scriptstyle \pm2.42}$ & $\mathbf{69.17}{\scriptstyle \pm1.12}$ & ${87.80}{\scriptstyle \pm2.36}$ & $74.96{\scriptstyle \pm0.66}$ & ${57.58}{\scriptstyle \pm1.23}$ \\
GSINA &$\mathbf{79.57}{\scriptstyle \pm1.38}$ % r=0.5, macro
&$67.86{\scriptstyle \pm0.91}$ % r=0.8
&$\mathbf{90.08}{\scriptstyle \pm2.06}$ % r=0.7
&$\mathbf{75.47}{\scriptstyle \pm0.55}$ % r=0.7
&$\mathbf{58.61}{\scriptstyle \pm1.09}$ \\% r=0.8
\bottomrule
\label{tab:gsat_ogbg}
\end{tabular}
% \end{sc}
}
\end{center}
%%\vspace{-30pt}
\end{table}%{wraptable}%
Based on the definition of our invariant subgraph in Eq.~\ref{eq:Gs}, the prediction process (the predictor $P_{\theta}(Y|G_S)$ in Eq.~\ref{eq:lo}) could be regularized by our GSINA message passing mechanism in Eq.~\ref{eq:msg_pass}.   
For the $l$-th GNN message passing layer, GSINA weights each message $m_{\theta}(\mathbf{h}_i^{(l)}, \mathbf{h}_j^{(l)}, \mathbf{h}_{ij}^{(l)})$ by edge attention $\alpha^E_{ij}$,  $\mathbf{h}_{ij}^{(l)}$ is the  representation of edge $(i, j)$ (if applicable), $\bigoplus$ is any permutation invariant aggregation function,  $\gamma_{\theta}$ is the GNN update function, and $\mathbf{h}_i^{(l+1)}$ is updated representation. If the graph representation $\mathbf{h}_G$ is obtained from a readout $f_{\theta}$ of node representations output from the $L$-th (final) GNN layer $\{\mathbf{h}_i^{(L)}|i\in\mathcal{V}\}$, each node representation $\mathbf{h}_i^{(L)}$ is weighted by our node attention $\alpha^V_i$ in our GSINA, GSINA is general due to its applicability to multiple-level (i.e. graph-level and node-level) GIL tasks:
% \begin{equation}
\begin{align}\label{eq:msg_pass}
    \mathbf{h}_i^{(l+1)} &= \gamma_{\theta}\left(\mathbf{h}_i^{(l)}, \bigoplus_{j \in \mathcal{N}_i} \alpha^E_{ij} * m_{\theta}(\mathbf{h}_i^{(l)}, \mathbf{h}_j^{(l)}, \mathbf{h}_{ij}^{(l)})\right), \\\label{eq:msg_pass2}
    \mathbf{h}_G\,\,\,\, &= f_{\theta}\left(\{\alpha^V_i * \mathbf{h}_i^{(L)} \vert i \in \mathcal{V}\}\right).
\end{align}
% \end{equation}

The procedure of training GSINA is shown in Alg.~\ref{alg:train}. In the next part, we will give the theoretical understanding of the convergence behavior of our approach.

%\yjc{Here we further discuss its connection and difference to other works.}

% \begin{figure}[tb!]
%     \centering
%     \includegraphics[width=0.45\textwidth]{vis/pipeline.png}
%     \caption{
%     %The 
% %two-stage 
% %pipeline of GSINA, first performing the subgraph extraction, then making a prediction based on the extracted subgraph.
% The proposed GSINA framework.
% }
%     %%\vspace{-20pt}
%     \label{fig:pipeline}
% \end{figure}
%{\color{blue}

\subsection{Graph Sinkhorn Attention: Theoretical Study on its Convergence Property}\label{sec:converge}
We present our theoretical results as follows.

\begin{proposition} \textbf{(exponential convergence of Algorithm~\ref{alg:train})} Given an initial matrix \( \mathbf{T}_0 \) and target row and column sums \( \mathbf{R} \) and \( \mathbf{C} \), respectively, the Sinkhorn algorithm alternates between row and column normalization steps, and the matrix \( \mathbf{T}_k \) converges to the optimal solution \( \mathbf{T}^\star \) of Eq.~\ref{eq:eot}, with exponential convergence speed. That is, there exists a constant \( \rho < 1 \) such that:
\[
\|\mathbf{T}_k - \mathbf{T}^\star\|_2 \leq \rho^k \|\mathbf{T}_0 - \mathbf{T}^\star\|_2,
\]
where \( \| \cdot \|_2 \) denotes the spectral norm of a matrix.
\end{proposition}
\begin{proof}
    To prove that the matrix \( \mathbf{T}_k \) converges to \( T^\star \) at exponential speed, we track the behavior of the error matrix \( \Delta_k = \mathbf{T}_k - \mathbf{T}^\star \) and show how its spectral norm decreases over iterations.
Define the error matrix as:
\[
\Delta_k = \mathbf{T}_k - \mathbf{T}^\star,
\]
where \( \mathbf{T}^\star \) is the limiting matrix that satisfies the row and column sum constraints:
\[
\mathbf{T}^\star \mathbf{1} =\mathbf{R} \quad \text{and} \quad \mathbf{T}^{\star\top} \mathbf{1} = \mathbf{R}.
\]

For each row normalization step, \( \mathbf{T}_k \) is updated as:
\[
\mathbf{T}_k = \operatorname{diag}\left(\mathbf{T}_{k-1}\mathbf{1} \oslash \mathbf{R}\right)^{-1} \mathbf{T}_{k-1}.
\]
This means that each row of \( \mathbf{T}_{k-1} \) is adjusted to match the target row sum \( \mathbf{R} \), reducing the row-wise error. Thus, the row normalization step reduces the error in the rows of \( \mathbf{T}_{k-1} \), as each row sum is made equal to the corresponding value in \( \mathbf{R} \). Similarly, at each column normalization step, the matrix \( \mathbf{T}_k \) is updated as:
\[
\mathbf{T}_k = \mathbf{T}_{k-1} \operatorname{diag}(\mathbf{T}_{k-1}^\top \mathbf{1} \oslash \mathbf{C}) ^{-1}.
\]
Here, each column of \( \mathbf{T}_{k-1} \) is adjusted to match the target column sum \( \mathbf{C} \), reducing the column-wise error. Hence, the column normalization step reduces the error in the columns of \( \mathbf{T}_{k-1} \), as each column sum is made equal to the corresponding value in \( \mathbf{C} \).

At each iteration, the error matrix \( \Delta_k \) is reduced by the row and column normalizations. Specifically, each of these normalization steps can be viewed as a contraction map that reduces the error proportionally. Thus, the error matrix \( \Delta_k \) continues to decrease with each iteration. Since both the row and column normalization steps are independent contractions, we can conclude that the error matrix \( \Delta_k \) will decay at an exponential rate, i.e. there exists a constant \( \rho < 1 \) such that:
\[
\|\Delta_k\|_2 \leq \rho \|\Delta_{k-1}\|_2.
\]

Because the error decays at a constant rate \( \rho \) per iteration, we can express the error matrix \( \Delta_k \) as:
\[
\|\Delta_k\|_2 \leq \rho^k \|\Delta_0\|_2.
\]
Thus, \( \mathbf{T}_k \) converges to \( \mathbf{T}^\star \) exponentially fast, with the error \( \Delta_k \) shrinking at an exponential rate.

This finishes the proof.

\end{proof}
%}
\section{Experiments}\label{sec:exp}
We conduct experiments on various tasks, including graph-level and node-level OOD generalizations. All experiments are conducted for 5 runs on RTX-2080Ti (11GB) GPUs, and the average and standard deviation are reported.

For graph-level GIL, we compare with both Information Bottleneck (IB) and top-$k$ based state-of-the-art methods, i.e. GSAT~\cite{miao2022interpretable} (graph learning via stochastic attention mechanism) and CIGA~\cite{chen2022learning} (causally invariant representations for out-of-distribution generalization on graphs), respectively. It is worth noting that the reason for performing separate experiments is for fair comparisons, as the experimental settings of these two benchmarks of GSAT and CIGA are quite different, including differences in datasets, GNN backbones, baselines, etc. Besides, both GSAT and CIGA involve extensive hyperparameter tuning. All these make the unified comparison nontrivial. %As will be shown in our evaluation, our separate experiments have indeed demonstrated the superiority of our methods over each of them and we have also released our source code. 

\begin{table*}[tb!]
	\centering
	\caption{Dataset statistics of the benchmark used in experiments of Table~\ref{tab:ciga_spm} and Table~\ref{tab:ciga}.}%\small\sc
 %%\vspace{-5pt}
	\label{tab:ciga-data}
	% \resizebox{\linewidth}{!}
 {
		\begin{small}
			\begin{tabular}{lccccccc}
				\toprule
				{Datasets} & {\# Training} & {\# Validation} & {\# Testing} & {\# Classes} & { \# Nodes} & { \# Edges}& {  Metrics}                                                                                                                           \\\midrule
				SPMotif           & $9,000$              & $3,000$                & $3,000$             & $3$                 & $44.96$            & $65.67$            & ACC     \\
    SST5              & $6,090$              & $1,186$                & $2,240$             & $5$                 & $19.85$            & $37.70$            & ACC     \\
				Twitter           & $3,238$              & $694$                  & $1,509$             & $3$                 & $21.10$            & $40.20$            & ACC     \\
				
				% CMNIST-sp         & $40,000$             & $5,000$                & $15,000$            & $2$                 & $56.90$            & $373.85$           & ACC     \\
				DrugOOD-Assay     & $34,179$             & $19,028$               & $19,032$            & $2$                 & $32.27$            & $70.25$            & ROC-AUC \\
				DrugOOD-Scaffold  & $21,519$             & $19,041$               & $19,048$            & $2$                 & $29.95$            & $64.86$            & ROC-AUC \\
				DrugOOD-Size      & $36,597$             & $17,660$               & $16,415$            & $2$                 & $30.73$            & $66.90$            & ROC-AUC \\
    PROTEINS          & $511$                & $56$                   & $112$               & $2$                 & $39.06$            & $145.63$           & MCC     \\
				DD                & $533$                & $59$                   & $118$               & 2                   & $284.32$           & $1,431.32$         & MCC     \\
				NCI1              & $1,942$              & $215$                  & $412$               & $2$                 & $29.87$            & $64.6$             & MCC     \\
				NCI109            & $1,872$              & $207$                  & $421$               & $2$                 & $29.68$            & $64.26$            & MCC     \\
				
				\bottomrule
			\end{tabular}	\end{small}}
\end{table*}

\begin{table*}[tb!]
	% %%\vspace{-0.4in}
	 \small
 %%\vspace{-5pt}
	\caption{Comparison with top-$k$ based GIL, graph-level OOD generalization on synthetic datasets used in the CIGA~\cite{chen2022learning} experiments: SPMotif-Struc and SPMotif-Mixed with varying hyperparameter bias $b$ for graph generation.}
	 %%\vspace{-10pt}
	\label{tab:ciga_spm}
	% \begin{sc}
		\begin{center}
			\begin{tabular}{lrrrrrr}
				\toprule       
    Methods &
				\multicolumn{3}{c}{SPMotif-Struc}   
    &
				\multicolumn{3}{c}{SPMotif-Mixed}        \\
    &
				\multicolumn{1}{c}{$b=0.33$}&
				\multicolumn{1}{c}{$b=0.60$}&
				\multicolumn{1}{c}{$b=0.90$}&
				\multicolumn{1}{c}{$b=0.33$}&
				\multicolumn{1}{c}{$b=0.60$}&
				\multicolumn{1}{c}{$b=0.90$}            \\
\midrule
				ERM~\cite{Vapnik_1991}                                    
    &$ 59.49 {\scriptstyle \pm3.50}$     
    &$ 55.48 {\scriptstyle \pm4.84}$
    &$ 49.64 {\scriptstyle \pm4.63}$
    &$ 58.18 {\scriptstyle \pm4.30}$            
    &$ 49.29 {\scriptstyle \pm8.17}$
    &$ 41.36 {\scriptstyle \pm3.29}$   
                                                 \\

			IRM~\cite{arjovsky2019invariant}                          
   &$ 57.15 {\scriptstyle \pm3.98}$            
   &$ 61.74 {\scriptstyle \pm1.32}$
   &$ 45.68 {\scriptstyle \pm4.88}$
   &$ 58.20 {\scriptstyle \pm1.97}$            
   &$ 49.29 {\scriptstyle \pm3.67}$
   &$ 40.73 {\scriptstyle \pm1.93}$                                               \\
				V-Rex~\cite{krueger2021out}                                       &$ 54.64 {\scriptstyle \pm3.05}$            
    &$ 53.60 {\scriptstyle \pm3.74}$
    &$ 48.86 {\scriptstyle \pm9.69}$
    &$ 57.82 {\scriptstyle \pm5.93}$            
    &$ 48.25 {\scriptstyle \pm2.79}$
    &$ 43.27 {\scriptstyle \pm1.32}$          
    \\
				EIIL~\cite{Creager_Jacobsen_Zemel_2020}                                        &$ 56.48 {\scriptstyle \pm2.56}$            
    &$ 60.07 {\scriptstyle \pm4.47}$
    &$ 55.79 {\scriptstyle \pm6.54}$
    &$ 53.91 {\scriptstyle \pm3.15}$            
    &$ 48.41 {\scriptstyle \pm5.53}$
    &$ 41.75 {\scriptstyle \pm4.97}$        
    \\
				IB-IRM~\cite{ahuja2021invariance}                                      &$ 58.30 {\scriptstyle \pm6.37}$            
    &$ 54.37 {\scriptstyle \pm7.35}$
    &$ 45.14 {\scriptstyle \pm4.07}$
    &$ 57.70 {\scriptstyle \pm2.11}$            
    &$ 50.83 {\scriptstyle \pm1.51}$
    &$ 40.27 {\scriptstyle \pm3.68}$   
                                                 \\

				CNC~\cite{Zhang_Sohoni_Zhang_Finn_R}                                         &$ 70.44 {\scriptstyle \pm2.55}$ 
    &$66.79 {\scriptstyle \pm9.42}$
    &$ 50.25 {\scriptstyle \pm10.7}$
    &$ 65.75 {\scriptstyle \pm4.35}$            
    &$ 59.27 {\scriptstyle \pm5.29}$
    &$ 41.58 {\scriptstyle \pm1.90}$                                               \\
				\midrule
    ASAP~\cite{ranjan2019asap}                                        
    &$ 64.87 {\scriptstyle \pm13.8}$            
    &$ 64.85 {\scriptstyle \pm10.6}$
    &$ 57.29 {\scriptstyle \pm14.5}$
    &$ 66.88 {\scriptstyle \pm15.0}$            
    &$ 59.78 {\scriptstyle \pm6.78}$
    &$ 50.45 {\scriptstyle \pm4.90}$  
    \\

				DIR~\cite{wu2022discovering}                                         &$ 58.73 {\scriptstyle \pm11.9}$   
    &$ 48.72 {\scriptstyle \pm14.8}$
    &$ 41.90 {\scriptstyle \pm9.39}$
    &$ 67.28 {\scriptstyle \pm4.06}$            
    &$ 51.66 {\scriptstyle \pm14.1}$
    &$ 38.58 {\scriptstyle \pm5.88}$          
    \\
				CIGAv1~\cite{chen2022learning}           
    &$ 71.07 {\scriptstyle \pm3.60}$   
    &$ 63.23 {\scriptstyle \pm9.61}$
    &$ 51.78 {\scriptstyle \pm7.29}$
    &$ 74.35 {\scriptstyle \pm1.85}$   
    &$ 64.54 {\scriptstyle \pm8.19}$
    &$ 49.01 {\scriptstyle \pm9.92}$                                         \\
				CIGAv2~\cite{chen2022learning}                          
    &$ \mathbf{77.33} {\scriptstyle \pm9.13}$   
    &$ 69.29 {\scriptstyle \pm3.06}$
    &$ 63.41 {\scriptstyle \pm7.38}$
    &$ 72.42 {\scriptstyle \pm4.80}$   
    &$ 70.83 {\scriptstyle \pm7.54}$
    &$54.25 {\scriptstyle \pm5.38}$   
    \\
    SuGAr~\cite{liu2025subgraph} & 72.85\,$\pm$\,$\scriptstyle 6.41$ & 70.01\,$\pm$\,$\scriptstyle 5.82$ & 67.74\,$\pm$\,$\scriptstyle 6.05$ & 78.10\,$\pm$\,$\scriptstyle 4.37$ & 73.42\,$\pm$\,$\scriptstyle 7.26$ & 64.71\,$\pm$\,$\scriptstyle 6.92$ \\
    IGM~\cite{jia2024graph} 
    & 74.10\,$\pm$\,$\scriptstyle 5.93$ 
    & 72.15\,$\pm$\,$\scriptstyle 4.27$ 
    & 69.17\,$\pm$\,$\scriptstyle 4.29$ 
    & 80.85\,$\pm$\,$\scriptstyle 4.12$ 
    & 75.91\,$\pm$\,$\scriptstyle 6.64$ 
    & 66.30\,$\pm$\,$\scriptstyle 7.83$
    \\
     GSINA (ours)                          
    & $75.49 {\scriptstyle \pm4.26}$
    &$\mathbf{74.25} {\scriptstyle \pm2.53}$
    &$\mathbf{73.54} {\scriptstyle \pm5.54}$
    & $\mathbf{82.70} {\scriptstyle \pm6.28}$
    &$\mathbf{77.03} {\scriptstyle \pm2.66}$
    &$\mathbf{68.89} {\scriptstyle \pm8.17}$
    \\
				\midrule
				Oracle (IID)      
    &$          88.70 {\scriptstyle \pm0.17}$               
    &$ 88.70 {\scriptstyle \pm0.17}$ 
    &$ 88.70 {\scriptstyle \pm0.17}$ 
    &$ 88.73 {\scriptstyle \pm0.25}$ 
    &$ 88.73 {\scriptstyle \pm0.25}$ 
    &$ 88.73 {\scriptstyle \pm0.25}$ 
    \\
				\bottomrule
				% \multicolumn{8}{l}{\rule{0pt}{8pt}$^\dagger$\text{\normalfont \small Higher accuracy and lower variance indicate better OOD generalization ability.}  }
			\end{tabular}
		\end{center}
  %%\vspace{-5pt}
	% \end{sc}	 %%\vspace{-10pt}
\end{table*}

For node-level GIL, we perform a comparison with the seminal EERM~\cite{wu2022handling} (Explore-to-Extrapolate Risk Minimization). Also we provide 1) hyperparameter studies, 2) ablation studies of GSINA components, 3) runtime and complexity analysis, and 4) interpretability analysis of the extracted invariant subgraph. We introduce the datasets, baselines, metrics and provide analysis. The source code is publicly available at \url{https://github.com/dingfangyu/GSINA}.
%All experiments are conducted for five runs on RTX-2080Ti (11GB) GPUs, and the average and standard deviation are both reported.

\subsection{Graph-Level Tasks: Compare with IB-Based GIL}
\label{sec:exp_gsat}
To make a direct and fair comparison with the IB-based GIL methods, we choose GSAT~\cite{miao2022interpretable} as the baseline. We perform evaluations strictly in line with its original experiment settings, including the datasets, GNN backbones, metrics, and optimization settings. To compare with GSAT, the hyperparameter $r$ for GSINA is chosen by validation. The statistics and evaluation metrics of these datasets are shown in Table~\ref{tab:gsat-data-1} and Table~\ref{tab:gsat-data-2}, and the corresponding results are given in Table~\ref{tab:gsat} and Table~\ref{tab:gsat_ogbg}.

\subsubsection{Datasets, Metrics and Baselines} 

\textbf{Datasets.} {
We use the synthetic Spurious-Motif (SPMotif) datasets from DIR~\cite{wu2022discovering}, where each graph is constructed by a combination of a motif graph directly determining the graph label, and a base graph providing spurious correlation to graph label, and we use datasets with spurious correlation degree $b=$ 0.5, 0.7 and 0.9, which serves as the hyperparameter for synthetic graph generation. For real-world datasets, we use MNIST-75sp~\cite{knyazev2019understanding}, where each image in MNIST  is converted to a superpixel graph, Graph-SST2~\cite{socher2013recursive,yuan2020explainability}, which is a sentiment analysis dataset, and each text sequence in SST2 is converted to a graph, following the splits in DIR~\cite{wu2022discovering}, Graph-SST2 contains degree shifts, and molecular property prediction datasets from the OGBG~\cite{wu2018moleculenet, hu2020ogb} benchmark (molhiv, molbace, molbbbp, molclintox, moltox21, molsider), as well as the DrugOOD dataset~\cite{drugOOD}.}

\textbf{Metrics.} We test the classification accuracy (ACC) for SPMotif datasets, MNIST-75sp, Graph-SST2, and ROC-AUC for OGBG  datasets following~\cite{miao2022interpretable}.

\textbf{Baselines.} Following the settings in GSAT~\cite{miao2022interpretable} and also the recent methods, we compare with interpretable GNNs namely, GIB~\cite{yu2020graph}, DIR~\cite{wu2022discovering}, IGM~\cite{jia2024graph}, and SuGAr~\cite{liu2025subgraph}. We use GIN~\cite{xu2018powerful} and PNA~\cite{corso2020principal} together as two optional backbones for GSAT and GSINA. We do not provide the comparison with the methods~\cite{ding2024sgood,sun2024dive} as they are not open-sourced when writing this article, and it is nontrivial to reproduce their implementation as there need additional technical detail. Besides, \cite{ding2024sgood} is designed for node classification only.

\subsubsection{Setup Details}
\textbf{GNN Backbone.}
For GSINA with GIN and PNA backbones for the experiments on GSAT benchmark (reported in Table~\ref{tab:gsat} and \ref{tab:gsat_ogbg}, respectively), our GNN backbone settings of GIN and PNA are strictly in line with those in the GSAT settings. We use 2 layers GIN with 64 hidden dimensions and 0.3 dropout ratio. There are 4 layers PNA with 80 hidden dimensions, with 0.3 dropout ratio, and no scalars are used. We directly follow PNA and GSAT using (mean, min, max, std) aggregators for OGBG datasets, and (mean, min, max, std, sum) aggregators for all other datasets in line with the literature. For all experiments of GSINA, the $\text{MLP}_{\phi}$ in our subgraph extractor $g_{\phi}$ is a 2-layer MLP  for simplicity, $\text{MLP}_{\phi}(\mathbf{n}_i,\mathbf{n}_j)$ is implemented by inputting concatenated $[\mathbf{n}_i,\mathbf{n}_j]$, and outputting a 1-dim edge score $s_{ij}$, then a batch normalization layer ($s := \frac{s - \text{mean}(s)}{\text{std}(s)}$) is applied to normalize the edge scores to a stable value range; and the aggregator of GSINA node attention to aggregate the edge attention values in a node's neighborhood is the `max' aggregator.

\textbf{Optimization.}
In line with GSAT, we use a batch size of 128 for all datasets. We use Adam~\cite{kingma2014adam} for graph classification reported in Table~\ref{tab:gsat} and Table~\ref{tab:gsat_ogbg} is strictly in line with the settings of GSAT. Our GSINA with GIN backbone uses 0.003 learning rate for Spurious-Motifs and 0.001 for all other datasets. Our GSINA with PNA backbone uses 0.003 learning rate for Graph-SST2 and Spurious-Motifs, and 0.001 learning rate for all other datasets. 

\textbf{Epoch.} {For graph classification on GSAT and CIGA, we perform early stop to avoid overfitting. Based on the difficulty of fitting the dataset, we set the early stopping patience to 10 for the SPMotif (used both in the experiments in GSAT and CIGA) datasets, DrugOOD-Size and TU datasets, 3 for Graph-SST2 and Graph-SST5, 5 for Twitter, {20 for DrugOOD-Assay/Scaffold} and 30 for MNIST-75sp. We do not use early stopping for OGBG datasets (molhiv, molbace, molbbbp, molclintox, moltox21, molsider), and simply train them to the end of epochs (200 epochs for GIN + GSINA, 100 for PNA + GSINA) for better performances. For TU datasets, like the practices of pre-training in CIGA, we pretrain them for 30  epochs to avoid underfitting. For EERM benchmark, we do not perform early stoppings nor pre-trainings, instead we follow the epochs used in EERM work and train to the end, which is 200 epochs for all datasets except for 500 epochs for OGB-Arxiv.}

\begin{table*}[tb!]
% \begin{wraptable}{r}{0.65\textwidth}
	% %%\vspace{-0.3in}
	% \scriptsize
 %%\vspace{-5pt}
 \centering
	\caption{Comparison with top-$k$ based GIL, graph-level OOD generalization on real-world datasets of  CIGA~\cite{chen2022learning} benchmark, \textbf{bold} font for the best performance on each dataset and \underline{underline} for the second.}
	\label{tab:ciga}
 	%%\vspace{-5pt}
	  \resizebox{\textwidth}{!}{
	% 	\begin{sc}
			\begin{tabular}{lccccccccc}
				\toprule
                    
                    % & \multicolumn{3}{c}{DrugOOD} 
                    % & \multicolumn{4}{c}{TU Datasets} 
                    % \\
                    Methods
                    & Graph-SST5  
                    & Twitter
                    & Drug-Assay
                    & Drug-Sca
                    & Drug-Size
                    & NCI1 
                    & NCI109 
                    & PROT 
                    & DD \\
				\midrule
				ERM~\cite{Vapnik_1991}  
                    & $43.89{\scriptstyle \pm1.73}$
                    & $60.81{\scriptstyle \pm2.05}$
                    & $71.79 {\scriptstyle \pm0.27}          $        
                    & $68.85 {\scriptstyle \pm0.62}$                 
                    & $66.70 {\scriptstyle \pm1.08}                  $
                    & $0.15 {\scriptstyle \pm0.05}              $
                    & $0.16 {\scriptstyle \pm0.02}                $
                    & $0.22 {\scriptstyle \pm0.09}                  $
                    & $0.27 {\scriptstyle \pm0.09} $                        \\
				
				IRM~\cite{arjovsky2019invariant}           
    & $43.69{\scriptstyle \pm1.26}$
    & $63.50{\scriptstyle \pm1.23}$  
    & $72.12 {\scriptstyle \pm0.49}      $         
    & $68.69 {\scriptstyle \pm0.65}     $         
    & $66.54 {\scriptstyle \pm0.42}   $        
    & $0.17 {\scriptstyle \pm0.02}        $    
    & $0.14 {\scriptstyle \pm0.01}  $           
    &$ 0.21 {\scriptstyle \pm0.09}       $       
    &$ 0.22 {\scriptstyle \pm0.08}     $                        \\
				V-Rex~\cite{krueger2021out}   & $43.28{\scriptstyle \pm0.52}$               
    & $63.21{\scriptstyle \pm1.57}$
    &$ 72.05 {\scriptstyle \pm1.25} $               
    &$ 68.92 {\scriptstyle \pm0.98}      $         
    &$ 66.33 {\scriptstyle \pm0.74}         $     
    & $0.15 {\scriptstyle \pm0.04}  $          
    & $0.15 {\scriptstyle \pm0.04}    $     
    & $0.22 {\scriptstyle \pm0.06}   $      
    & $0.21 {\scriptstyle \pm0.07}             $               \\
				EIIL~\cite{Creager_Jacobsen_Zemel_2020}   
    & $42.98{\scriptstyle \pm1.03}$
    & $62.76{\scriptstyle \pm1.72}$          
    & $72.60 {\scriptstyle \pm0.47}$               
    &$ 68.45 {\scriptstyle \pm0.53}         $    
    &$ 66.38 {\scriptstyle \pm0.66}    $       
    & $0.14 {\scriptstyle \pm0.03}      $  
    &$ 0.16 {\scriptstyle \pm0.02}     $  
    &$ 0.20 {\scriptstyle \pm0.05}     $    
    &$ 0.23 {\scriptstyle \pm0.10}      $                       \\
				IB-IRM~\cite{ahuja2021invariance}& $40.85{\scriptstyle \pm2.08}$               
    & $61.26{\scriptstyle \pm1.20}$       
    &$ 72.50 {\scriptstyle \pm0.49}         $       
    & $68.50 {\scriptstyle \pm0.40}          $    
    & $66.64 {\scriptstyle \pm0.28}    $       
    & $0.12 {\scriptstyle \pm0.04}     $    
    & $0.15 {\scriptstyle \pm0.06}      $   
    & $0.21 {\scriptstyle \pm0.06}    $        
    & $0.15 {\scriptstyle \pm0.13}          $                    \\
				CNC~\cite{Zhang_Sohoni_Zhang_Finn_R}       
    & $42.78{\scriptstyle \pm1.53}$
    & $61.03{\scriptstyle \pm2.49}$              
    &$ 72.40 {\scriptstyle \pm0.46}  $             
    &$ 67.24 {\scriptstyle \pm0.90}   $            
    & $65.79 {\scriptstyle \pm0.80} $             
    & $0.16 {\scriptstyle \pm0.04}  $         
    &$ 0.16 {\scriptstyle \pm0.04}  $          
    &$ 0.19 {\scriptstyle \pm0.08}     $       
    &$ 0.27 {\scriptstyle \pm0.13}   $                        \\
    \midrule
   ASAP~\cite{ranjan2019asap}      
   & $44.16{\scriptstyle \pm1.36}$
   & $60.68{\scriptstyle \pm2.10}$
    & $70.51 {\scriptstyle \pm1.93}                   $
    & $66.19 {\scriptstyle \pm0.94}              $
    & $64.12 {\scriptstyle \pm0.67}        $
    & $0.16 {\scriptstyle \pm0.10}         $
    & $0.15 {\scriptstyle \pm0.07}            $
    & $0.22 {\scriptstyle \pm0.16}            $
    & $0.21 {\scriptstyle \pm0.08}    $                         \\
				GIB~\cite{yu2020graph}        & $38.64{\scriptstyle \pm4.52}$             
    & $48.08{\scriptstyle \pm2.27}$
    & $63.01 {\scriptstyle \pm1.16} $                
    & $62.01 {\scriptstyle \pm1.41}     $         
    & $55.50 {\scriptstyle \pm1.42}    $          
    &$ 0.13 {\scriptstyle \pm0.10}      $      
    & $0.16 {\scriptstyle \pm0.02}     $          
    & $0.19 {\scriptstyle \pm0.08}    $           
    & $0.01 {\scriptstyle \pm0.18}     $                      \\
				DIR~\cite{wu2022discovering}  & $41.12{\scriptstyle \pm1.96}$         
    & $59.85{\scriptstyle \pm2.98}$      
    &$ 68.25 {\scriptstyle \pm1.40}    $            
    &$ 63.91 {\scriptstyle \pm1.36}        $      
    &$ 60.40 {\scriptstyle \pm1.42}  $                
    &$ 0.21 {\scriptstyle \pm0.06}   $        
    &$ 0.13 {\scriptstyle \pm0.05}    $      
    &$ 0.25 {\scriptstyle \pm0.14}   $          
    &$ 0.20 {\scriptstyle \pm0.10}  $                          \\
    % \midrule
	CIGAv1~\cite{chen2022learning}  
 & $44.71{\scriptstyle \pm1.14}$
 & $63.66{\scriptstyle \pm0.84}$ 
 & ${72.71 }{\scriptstyle \pm0.52}  $  
 & $69.04 {\scriptstyle \pm0.86}   $
 & $67.24 {\scriptstyle \pm0.88} $
 & ${0.22} {\scriptstyle \pm0.07}$            
 &$ \underline{0.23} {\scriptstyle \pm0.09}$    
 &$ \underline{0.40}{\scriptstyle \pm0.06}     $
 &$\mathbf{0.29}{\scriptstyle \pm0.08}  $           \\
				CIGAv2~\cite{chen2022learning}& $45.25{\scriptstyle \pm1.27}$         
    & $\underline{64.45}{\scriptstyle \pm1.99}$
    &$ \mathbf{73.17} {\scriptstyle \pm0.39}  $   
    &$ \mathbf{69.70} {\scriptstyle \pm0.27}   $  
    & $\mathbf{67.78} {\scriptstyle \pm0.76}  $
    &$ 0.27 {\scriptstyle \pm0.07}  $
    & $0.22 {\scriptstyle \pm0.05}     $       
    &$ 0.31 {\scriptstyle \pm0.12} $     
    &$ 0.26 {\scriptstyle \pm0.08}   $                     \\
    SuGAr~\cite{liu2025subgraph} & 45.30\,$\pm$\,$\scriptstyle 1.42$ & 63.91\,$\pm$\,$\scriptstyle 2.25$ & 72.62\,$\pm$\,$\scriptstyle 0.53$ & 69.01\,$\pm$\,$\scriptstyle 0.44$ & 67.01\,$\pm$\,$\scriptstyle 0.72$ & 0.27\,$\pm$\,$\scriptstyle 0.06$ & 0.26\,$\pm$\,$\scriptstyle 0.08$ & 0.40\,$\pm$\,$\scriptstyle 0.07$ & 0.29\,$\pm$\,$\scriptstyle 0.06$ \\
    IGM~\cite{jia2024graph} & \underline{45.70}\,$\pm$\,$\scriptstyle 1.56$ 
    & 64.40\,$\pm$\,$\scriptstyle 2.17$ & \underline{73.16}\,$\pm$\,$\scriptstyle 0.55$ & 69.25\,$\pm$\,$\scriptstyle 0.71$ & 67.38\,$\pm$\,$\scriptstyle 0.79$ & 0.28\,$\pm$\,$\scriptstyle 0.04$ & $\mathbf{0.27}$\,$\pm$\,$\scriptstyle 0.06$ & $\mathbf{0.41}$\,$\pm$\,$\scriptstyle 0.07$ & $\mathbf{0.29}$\,$\pm$\,$\scriptstyle 0.06$ \\

    GSINA (ours) 
    & $\mathbf{45.84}{\scriptstyle \pm0.52} $ % macro 3
    & $\mathbf{64.64}{\scriptstyle \pm1.71}$
    &$72.84 {\scriptstyle \pm0.50}$ % macro  20, 20, 0, 0
    & $\underline{69.57}{\scriptstyle \pm 0.39}$ % macro 20, 20, 0, 0
    &$\underline{ 67.48} {\scriptstyle \pm0.33} $ %  macro, 20 10,0,0
    &$\mathbf{ 0.28} {\scriptstyle \pm0.07} $ % pretrain 20/30, pat 10, 0,0, macro
    & ${ 0.21} {\scriptstyle \pm0.04} $           % macro ,pre 30, pat 10, 0,0
    &$\mathbf{ 0.41} {\scriptstyle \pm0.07}     $ % pretrain 30, patience 10, a=0, b=0
    &$\mathbf{ 0.29} {\scriptstyle \pm0.07}$  \\% pretrain 30, patience 10, a=0, b=0
    \midrule
    Oracle (IID)
    & $48.18{\scriptstyle \pm1.00}$
    & $64.21{\scriptstyle \pm1.77}$
    & $85.56 {\scriptstyle \pm1.44}       $
    &$ 84.71 {\scriptstyle \pm1.60}          $                       
    &$85.83 {\scriptstyle \pm1.31} $
    &$ 0.32 {\scriptstyle \pm0.05} $
    &$ 0.37 {\scriptstyle \pm0.06}  $
    & $0.39 {\scriptstyle \pm0.09}   $
    &$ 0.33 {\scriptstyle \pm0.05}$                   \\
				\bottomrule
			\end{tabular}
% 		\end{sc}
  }
%%\vspace{-20pt}
	% %%\vspace{-0.15in}
% \end{wraptable}
\end{table*}

\textbf{Settings of the Subgraph Ratio Hyperparameter $r$.}
% 925664, 786135, 58778
For GIN backboned GSINA, we set $r=$ 0.9 for OGBG-Molhiv, 0.2 for Graph-SST2, 0.5 for MNIST-75sp, 0.6 for SPMotif-0.5, 0.6 for SPMotif-0.7, and 0.4 for SPMotif-0.9. For PNA backboned GSINA, we set $r=$ 0.7 for OGBG-Molhiv, 0.5 for OGBG-Molbace, 0.8 for OGBG-Molbbbp, 0.7 for OGBG-Molclintox, 0.7 for OGBG-Moltox21, 0.8 for OGBG-Molsider, 0.8 for Graph-SST2, 0.6 for MNIST-75sp, 0.1 for SPMotif-0.5, 0.3 for SPMotif-0.7, and 0.5 for SPMotif-0.9.
%{We perform $r$ selection based on the validation performances. 

\textbf{Performances Analysis.} 
Table~\ref{tab:gsat} reports the graph classification performances on GSAT benchmark. GSINA achieves better performances than the baselines of interpretable GNNs GIB, DIR, and GSAT. In particular, GSINA outperforms GSAT by large margins on all 6 datasets for both GNN backbones GIN and PNA. Table~\ref{tab:gsat_ogbg} reports the graph classification performances for another 5 OGBG-Mol datasets with smaller sizes than those in Table~\ref{tab:gsat}. Following GSAT, we compare with PNA backboned baselines, where GSINA also mostly outperforms.

These improvements show the effectiveness of our separability compared with GSAT (based on IB constraint). Especially, we observe relatively big improvements on SPMotif datasets. There are about at most 10\% improvements than GSAT in classification accuracy on SPMotif ($b=0.7$) in Table~\ref{tab:gsat}. It indicates the superiority of GSINA on the datasets with more distinguishable subgraph properties, such as the SPMotif datasets, where the invariant subgraphs can be clearly separable due to their data generation processes.
Notably, GSINA also consistently outperforms recent baselines IGM and SuGAr across all bias levels, further validating the effectiveness of our soft, separable, and fully differentiable subgraph extraction framework.

\subsection{Graph-Level Tasks: Compare with Top-$k$ Based GIL}
To make a comparison with the top-$k$ based GIL methods, we choose CIGA~\cite{chen2022learning} as the baseline. The statistics and evaluation metrics of these datasets are shown in Table~\ref{tab:ciga-data} and the experiment results are given in Table~\ref{tab:ciga_spm} and Table~\ref{tab:ciga}.

% For direct and fair comparisons with the two GIL SOTAs, GSAT~\cite{miao2022interpretable} and CIGA~\cite{chen2022learning}, each respectively using different datasets and GNN backbones, we perform evaluations strictly in line with their original corresponding settings and the results are given in Table~\ref{tab:gsat},~\ref{tab:gsat_ogbg} and Table~\ref{tab:ciga_spm},~\ref{tab:ciga} respectively. 
% To compare with GSAT, the hyperparameter $r$ for GSINA is chosen according to the validation performances. 
% For CIGA, as CIGA also performs top-$r$ subgraph extractions, $r$ follows the settings of CIGA on the datasets in Table~\ref{tab:ciga_spm},~\ref{tab:ciga}.

\subsubsection{Datasets, Metrics and Baselines} 

\textbf{Datasets.} {We use the synthetic SPMotif datasets from DIR, with structural shift degrees $b=$ 0.33, 0.6 and 0.9, denoted as SPMotif (-struc). Besides, we use the SPMotif (-mixed) from CIGA~\cite{chen2022learning}, whose distribution shifts are additionally mixed with attribute shifts. 
For real-world datasets, in line with CIGA, to validate the generalization performance with more complicated relationships under distribution shifts, we use sentiment analysis datasets Graph-SST5~\cite{yuan2020explainability} and Twitter~\cite{Dong_Wei_Tan_Tang_Zhou_Xu_2014} with degree shifts, DrugOOD datasets~\cite{drugOOD}, which is from AI-aided Drug Discovery, the split schemes including assay, scaffold and size, and the datasets from the TU~\cite{morris2020tudataset} benchmarks (nci1, nci109, proteins, dd) to examine the OOD generalization under graph size shifts.}

\textbf{Metrics.} {We test the classification accuracy (ACC) for SPMotif datasets, Graph-SST5, Twitter, ROC-AUC for DrugOOD datasets, and Matthews correlation coefficient (MCC) for TU datasets  following~\cite{bevilacqua2021size,chen2022learning}. 
}

\textbf{Baselines.} 
Following the experimental protocol in CIGA~\cite{chen2022learning}, in addition to ERM~\cite{Vapnik_1991}, we compare with interpretable GNNs GIB, DIR, ASAP Pooling~\cite{ranjan2019asap}, and the invariant learning methods IRM~\cite{arjovsky2019invariant}, V-Rex~\cite{krueger2021out}, IB-IRM~\cite{ahuja2021invariance}, EIIL~\cite{Creager_Jacobsen_Zemel_2020}, CNC~\cite{Zhang_Sohoni_Zhang_Finn_R}. The Oracle (IID) performances without distribution shifts are also reported. We use the GNN architectures in line with CIGA.
\subsubsection{Setup Details}
\textbf{GNN Backbone Architectures.}
{For GSINA with GCN or GIN layers for the experiments on CIGA benchmark (reported in Table~\ref{tab:ciga_spm} and Table~\ref{tab:ciga}), our GNN backbone settings  are also strictly in line with those in CIGA settings. We use 3-layer GNN with Batch Normalization between layers and JK residual connections at last layer. We use GCN with mean readout  for all datasets except the Proteins and DrugOOD datasets. For Proteins, we use GIN and max readout. For DrugOOD datasets, we use 4-layer GIN with sum readout. The hidden dimensions are fixed as 32 for SPMotif, TU datasets, and 128 for SST5, Twitter, and DrugOOD datasets.
}

\textbf{Batch size.} In line with CIGA, we use batch size $B=32$ for all datasets, except for DrugOOD, where we set $B=128$.

\textbf{Optimizer.}
{We use the Adam~\cite{kingma2014adam} for graph classification reported in Table~\ref{tab:ciga_spm} and Table~\ref{tab:ciga} in line with the settings of CIGA. We set the learning rate 0.001.
 }

\begin{figure*}[tb!]
    \centering
    \includegraphics[width=0.9\linewidth]{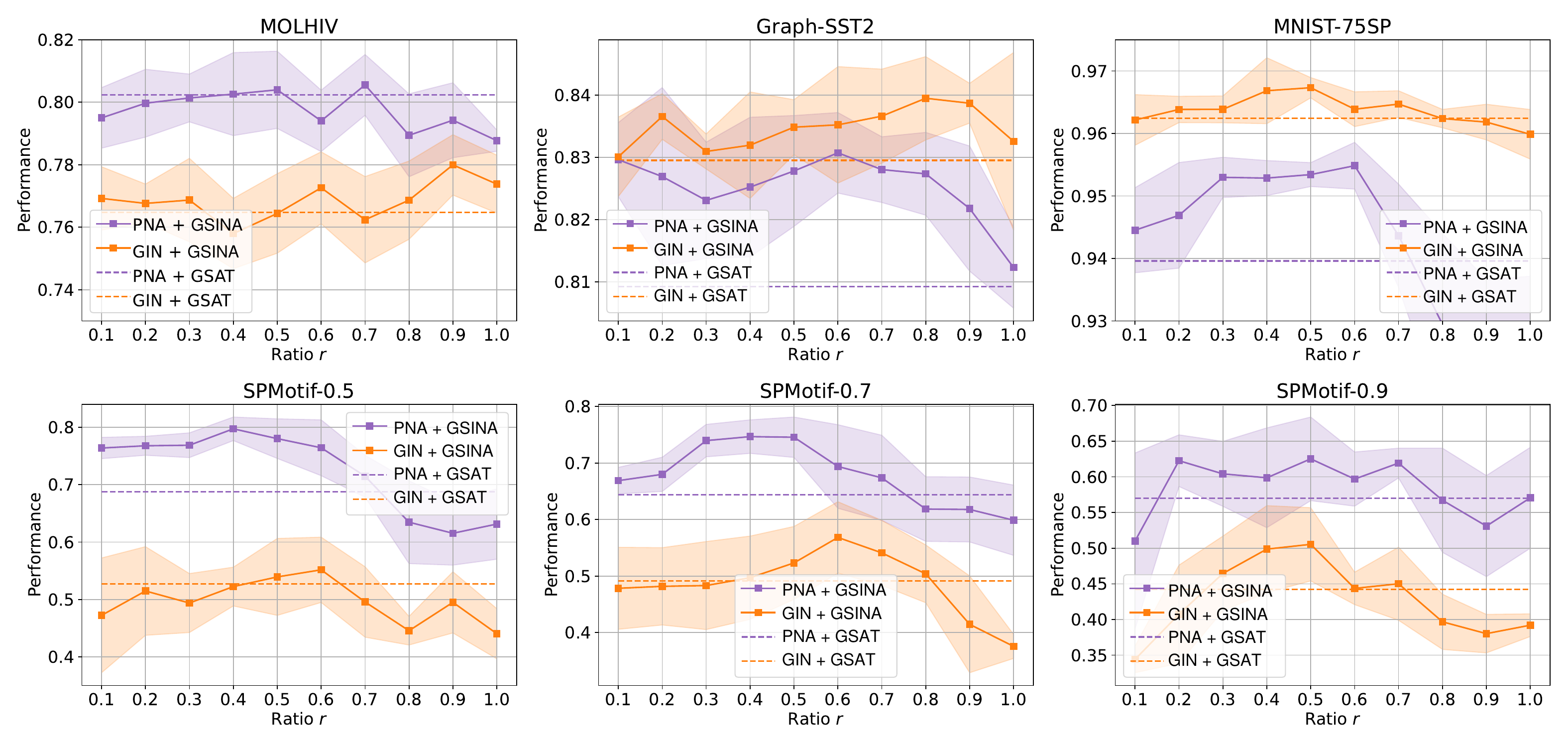}%%\vspace{-8pt}
    \caption{Hyperparameter sensitivity (classification accuracy as y-axis metric) analysis of the separability ratio $r$ in GSINA.}
    \label{fig:test_r}
%%\vspace{-20pt}
\end{figure*}

\begin{figure*}[tb!]
    \centering
    %%\vspace{-5pt}
    \includegraphics[width=0.92\linewidth]{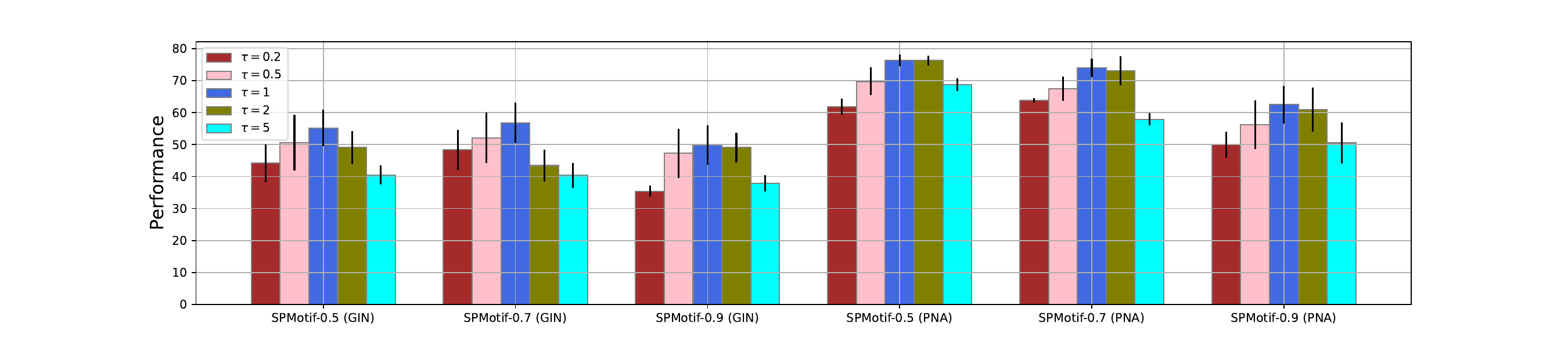}
    %%\vspace{-20pt}
    \caption{Hyperparameter sensitivity (classification accuracy as y-axis metric) of the temperature $\tau$ (for softness) in GSINA.}
    %%\vspace{-20pt}
    \label{fig:tau}
\end{figure*}

\textbf{Epoch.} {
We perform early stopping to avoid overfitting. Based on the difficulty of fitting the dataset, we set the  early stopping patience to 10 for the SPMotif datasets, DrugOOD-Size and TU datasets, 3 for Graph-SST5, 5 for Twitter, and 20 for DrugOOD-Assay/Scaffold. For the TU datasets, like the practices of pre-training in CIGA, we pretrain them for 30 epochs to avoid underfitting. }

\textbf{Hyperparameter settings of subgraph ratio $r$.}
% 925664, 786135, 58778
{As CIGA also performs top-$r$ subgraph extractions, we regard its settings of $r$ as reasonable and follow them: 0.25 for SPMotif, 0.3 for Proteins and DD, 0.6 for NCI1, 0.7 for NCI109, 0.5 for SST5 and Twitter, and 0.8 for DrugOOD, respectively.}

\textbf{Performances analysis.} 
Table~\ref{tab:ciga_spm} and and Table~\ref{tab:ciga} report the OOD generalization performances on the benchmarks used in the CIGA work. GSINA outperforms all the baselines of interpretable GNNs (ASAP, GIB, and DIR) by large margins. For the invariant learning baselines, GSINA also achieves better performances. Compared with CIGA, it achieves competitive performances on SPMotif (except for -struct, $b=0.33$), Graph-SST5, Twitter, proteins and dd. On nci1, nci109, and DrugOOD. GSINA produces results comparable to CIGA, indicating the difficulties of these OOD generalization tasks.
%meanwhile, the improvements of CIGA compared to ERM on DrugOOD are also by little margin.
% \hzp{Compared with CIGA, GSINA outperforms across various benchmarks such as SPMotif (except for -struct, $b=$ 0.33), Graph-SST5, Twitter, proteins, and dd. On nci1, nci109, and DrugOOD datasets, GSINA produces results comparable to CIGA, indicating the difficulties of these OOD generalization tasks. Notably, the marginal improvements of CIGA over ERM on the DrugOOD datasets are also relatively minor.}

These improvements show the effectiveness of our softness and fully differentiability computing compared with CIGA (based on top-$k$). There are also large improvements on SPMotif datasets. 
Similar to Sec.~\ref{sec:exp_gsat}, about 15\% improvement over CIGA in classification accuracy on SPMotif (-mixed, the hyperparameter $b=0.9$) in Table~\ref{tab:ciga_spm} is observed.  

% Moreover, there are cases (on Twitter, prot) where our GSINA outperforms the in-distribution oracle (by ERM, in line with the CIGA experiment design), which demonstrates our GSINA could find helpful crucial subgraphs for graph label prediction. 

\subsection{Hyperparameter Studies for Graph-level Tasks}\label{para:hp}
As shown in Fig.~\ref{fig:test_r}, for the six datasets used in the GSAT benchmark and reported in Table~\ref{tab:gsat}, there are rough `increasing-decreasing' patterns in the curve of predictive performance and subgraph ratio (i.e. separability) $r$, which reflects GSINA is sensible to the hyperparameter $r$. The patterns of the curves in Fig.~\ref{fig:test_r} show that a too-small or a too-large $r$ results in worse generalization. When $r$ is too small, it is more likely that the extracted subgraph $G_S$ is too sparse and lacks information; when $r$ is too large, it also results in a $G_S$ lack of information, as more redundant parts of the input graph $G$ would be selected, which results in a $G_S$ with little difference to $G$. When $r=1.0$, GSINA degenerates to ERM with all edge attention being set to 1.

\begin{figure*}[t]
  \centering
  % ---- Row 1 ----
  \begin{subfigure}[t]{0.37\textwidth}
    \centering
    \includegraphics[width=\linewidth]{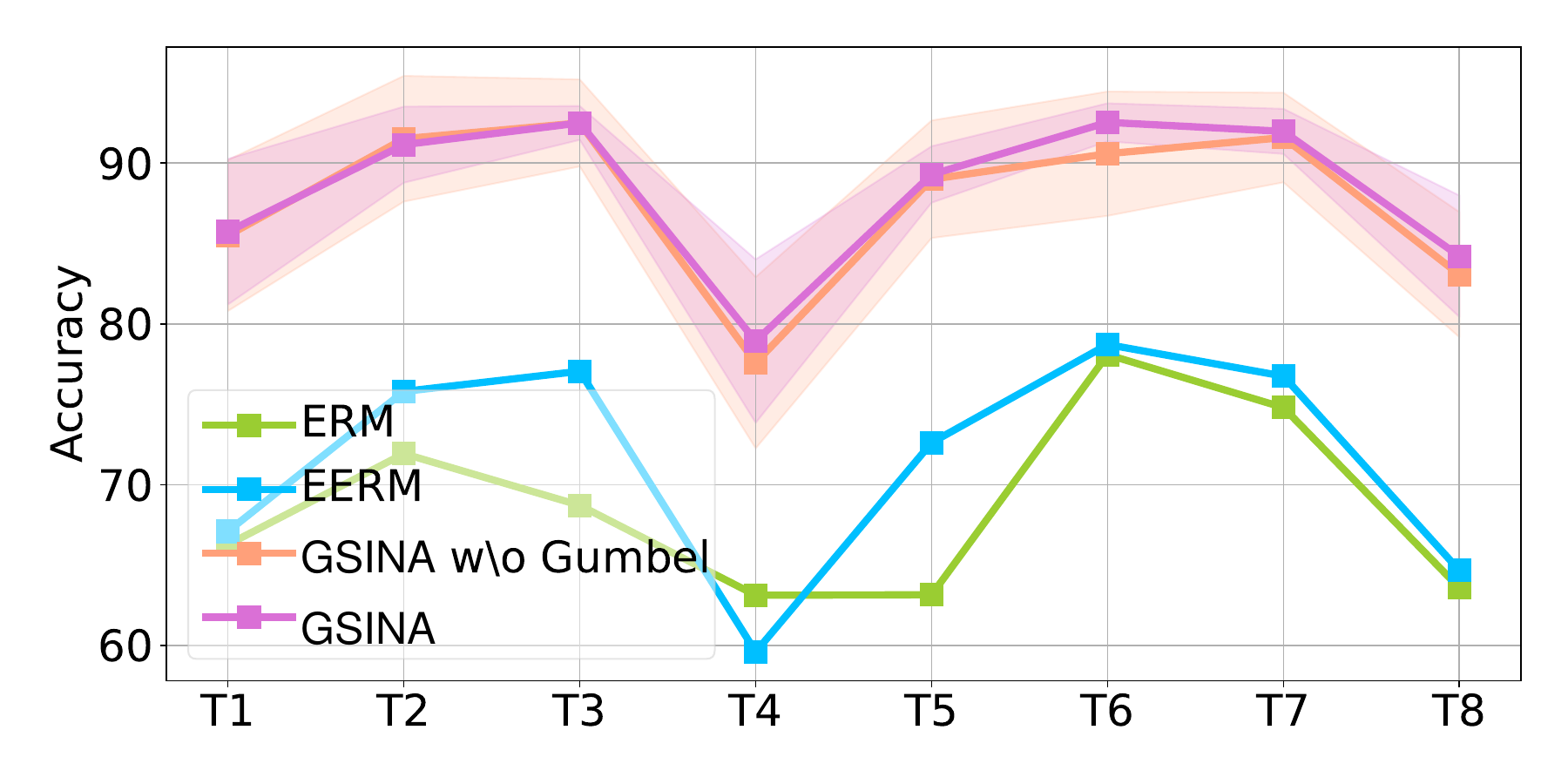}
    \caption{Cora}
  \end{subfigure}\hfill
  \begin{subfigure}[t]{0.37\textwidth}
    \centering
    \includegraphics[width=\linewidth]{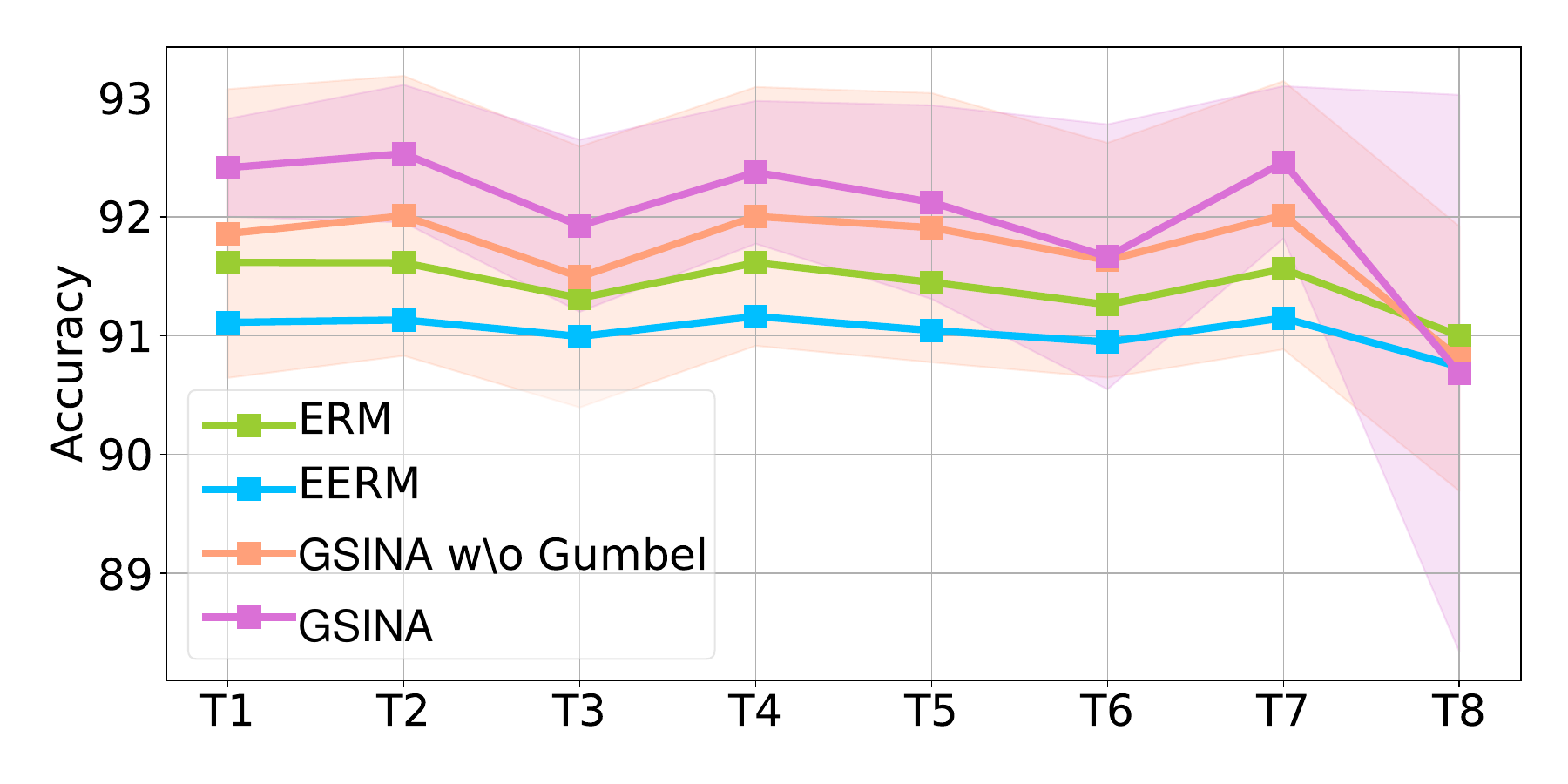}
    \caption{Amazon-Photo}
  \end{subfigure}\hfill
  \begin{subfigure}[t]{0.25\textwidth}
    \centering
    \includegraphics[width=\linewidth]{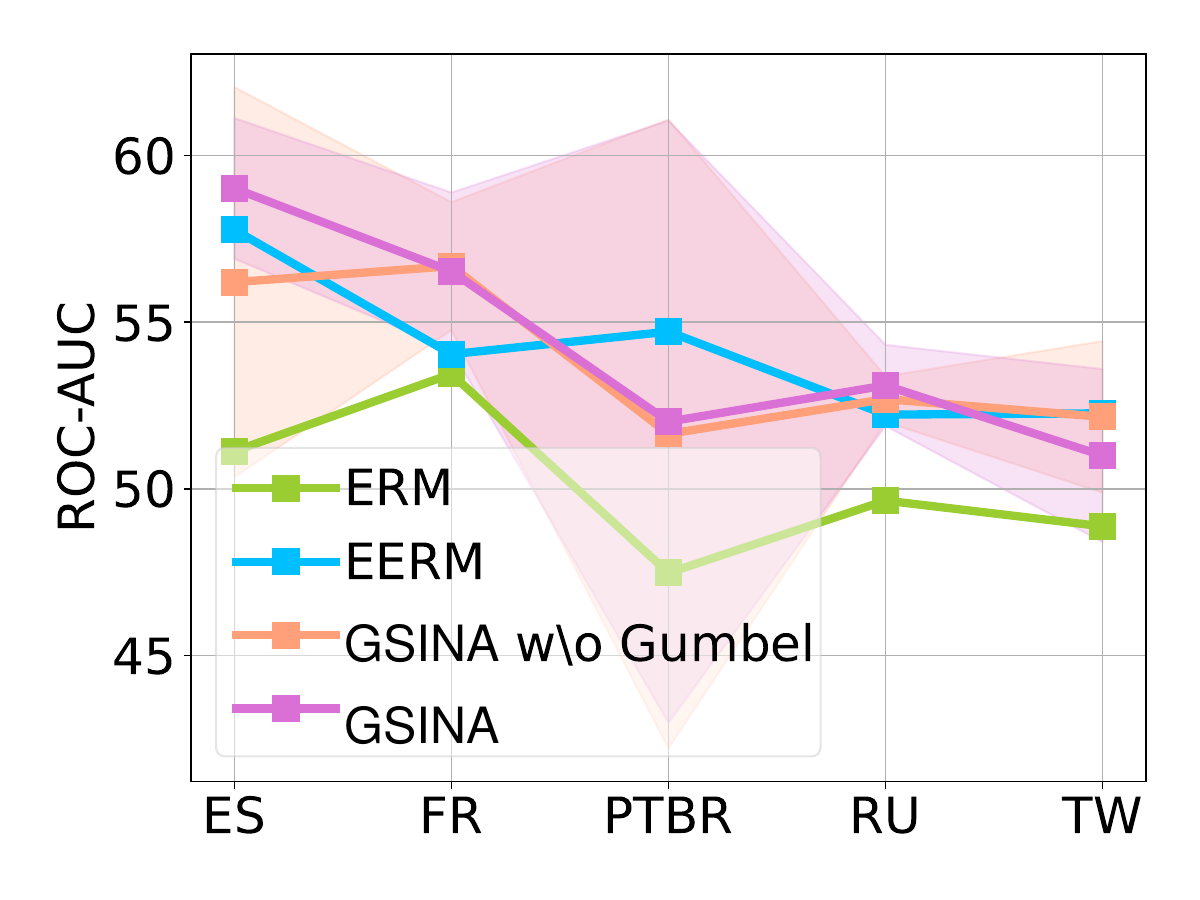}
    \caption{Twitch-explicit}
  \end{subfigure}

  \medskip

  % ---- Row 2 ----
  \begin{subfigure}[t]{0.38\textwidth}
    \centering
    \includegraphics[width=\linewidth]{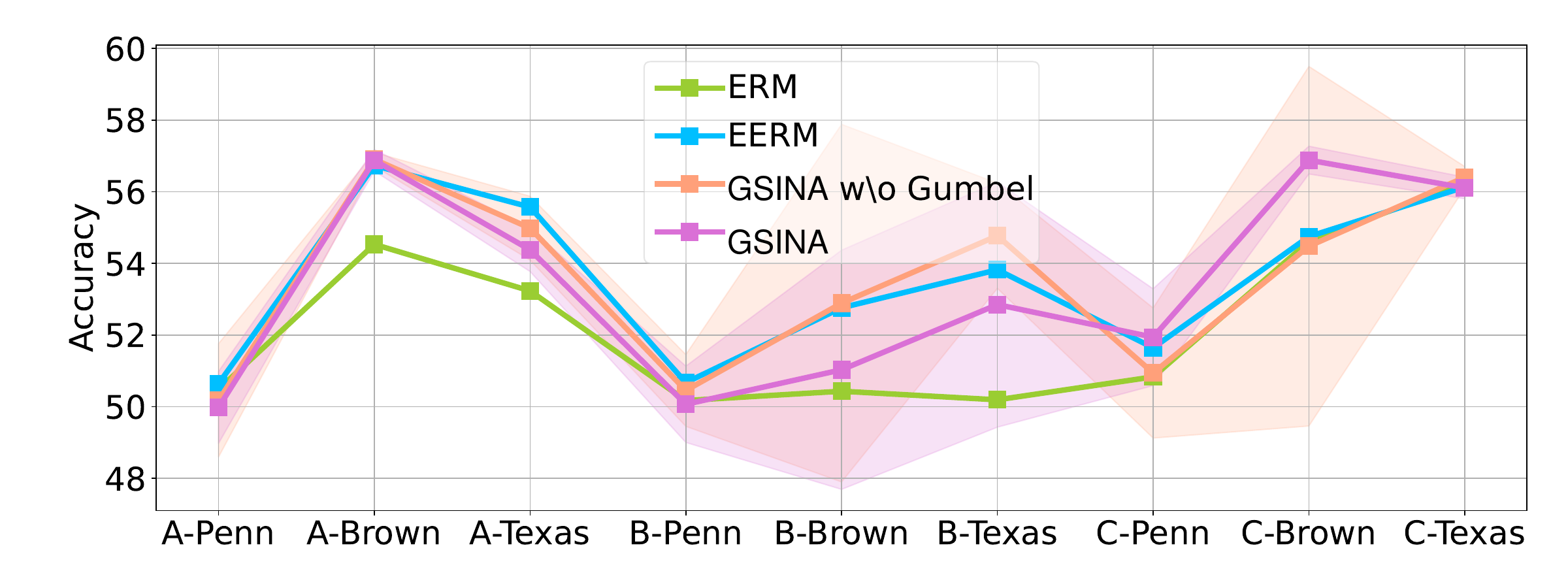}
    \caption{Facebook-100}
  \end{subfigure}\hfill
  \begin{subfigure}[t]{0.38\textwidth}
    \centering
    \includegraphics[width=\linewidth]{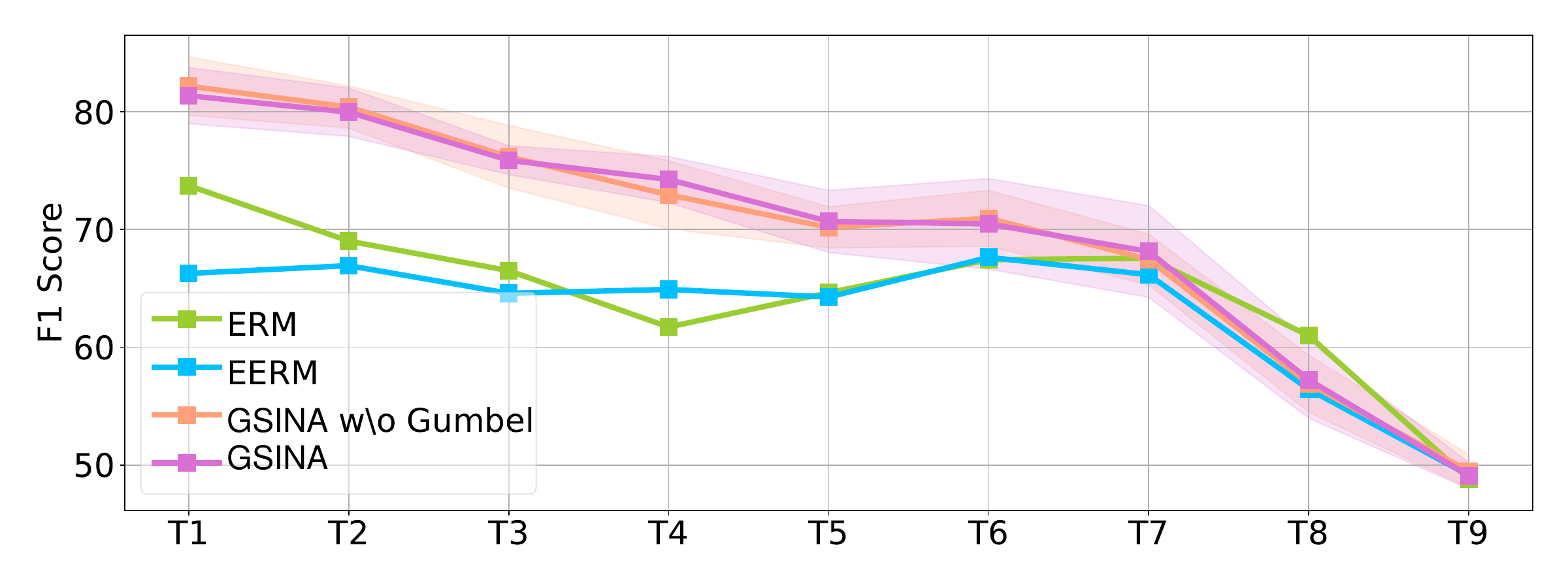}
    \caption{Elliptic}
  \end{subfigure}\hfill
  \begin{subfigure}[t]{0.23\textwidth}
    \centering
    \includegraphics[width=\linewidth]{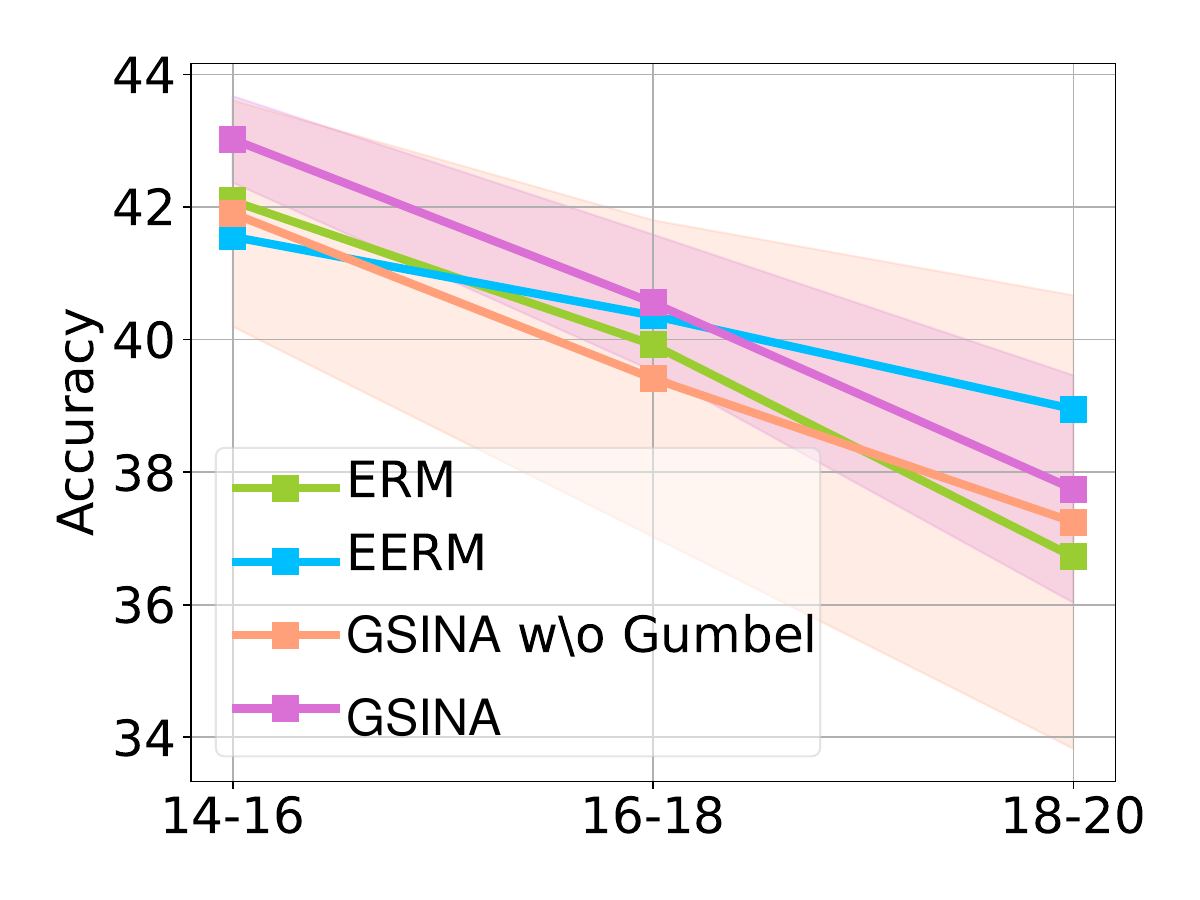}
    \caption{OGB-Arxiv}
  \end{subfigure}

  \caption{Node-level OOD generalization performances across three types of distribution shift. 
  Panels a and b correspond to \textbf{Artificial Transformation}; 
  panels c and d correspond to \textbf{Cross-Domain Transfers}; 
  panels e and f correspond to \textbf{Temporal Evolution}. The x-axis represent different datasets.  
  The shaded area denotes the variance.}
  \label{fig:node_ood}
\end{figure*}

As shown in Fig.~\ref{fig:tau}, we analyze the hyperparameter sensitivity of the Sinkhorn temperature $\tau \in \{0.2, 0.5, 1,2,5\}$ on the SPMotif datasets (of GSAT benchmark, with both GIN and PNA backboned GSINA), which demonstrates that $\tau=1$ is a reasonable and effective choice. When $\tau$ is too small, the model lacks softness and generates too `hard' subgraphs; when $\tau$ is too large, the model is too soft and generates subgraphs that are too smooth and lack information. Both cases lead to reduced performances.

%\subsubsection{More Hyperparameter Settings}
Note that in GSINA, we globally fix the Sinkhorn temperature $\tau$ (for softness) and Gumbel noise factor $\sigma$ (for randomness) to $1$, and the Sinkhorn iteration count to $10$.

Last but not least, we show that GSINA's model selection procedure is simpler than CIGA (the `hard' top-$k$ based GIL SOTA). Specifically, with other settings fixed, what is required to tune in GSINA is just the separability $r$, while CIGA has additional loss balancing hyperparameters (2 losses corresponding to CIGAv1 and CIGAv2) to tune.

\subsection{Experiments on Node-Level GIL Tasks}
\subsubsection{Datasets, Metrics and Baselines} 
We follow the datasets and protocols used in EERM~\cite{wu2022handling}, which involve three types of distribution shifts: 
1) ``Artificial Transformation'': synthetic spurious features are added to Cora and Amazon-Photo, and there are 8 testing graphs (T1 $\sim$ T8) for both datasets, 
2) ``Cross-Domain Transfers'': each graph in Twitch-explicit and Facebook-100 corresponds to distinct domains; for Twitch-explicit, DE is used for training, ENGB for validation and 5 graphs ES, FR, PTBR, RU, TW for testing;  for Facebook-100, 3 different training sets are used,  we denote them as A = (Johns Hopkins, Caltech, Amherst), B = (Bingham, Duke, Princeton), C = (WashU, Brandeis, Carnegie), the validation set is (Cornell, Yale), and the testing set is (Penn, Brown, Texas), 
3) ``Temporal Evolution'': train/val/test splits for Elliptic and OGB-Arxiv are made by time, Elliptic provides 9 test graphs (T1 $\sim$ T9), and OGB-Arxiv provides 3 time windows (14-16, 16-18, 18-20). The dataset details are shown in Table~\ref{tab:eerm-data}.

We test the node classification accuracy (ACC) on Cora, Amazon-Photo, Facebook-100, OGB-Arxiv, ROC-AUC for Twitch-explicit, and F1-score for Elliptic.

We compare with ERM~\cite{Vapnik_1991} and EERM~\cite{wu2022handling}, following the settings in~\cite{wu2022handling}, we use GCN~\cite{kipf2016semi} as backbone subgraph extractors, predictors and spurious features generators (if applicable) for Cora, Amazon-Photo,  Twitch-explicit and Facebook-100, SAGE~\cite{hamilton2017inductive} for Elliptic and OGB-Arxiv.

\subsubsection{Setup Details}
\textbf{GNN Backbone Architectures.}{
For GSINA with GCN and SAGE backbones for the experiments on EERM benchmark (in 
% Fig.~\ref{fig:node1},~\ref{fig:node2},~\ref{fig:node3}
Fig.~\ref{fig:node_ood}
), our GNN backbone settings of GCN and SAGE are strictly in line with those in EERM settings, where we use ReLU as the activation,  add self-loops and use batch normalization for graph convolution in each layer. We use 2 layers GCN with hidden size 32  for Cora, Amazon-Photo,  Twitch-explicit,  and Facebook-100;  and 5 layers SAGE with hidden size 32  for Elliptic and OGB-Arxiv.
}

\textbf{Optimization.}
{For all datasets, we follow the AdamW optimizers used in EERM. To reproduce EERM results, we strictly follow its original settings. For GSINA, we use the same settings with the experiments of ERM, where we use 0.01 learning rate for Cora, Amazon-Photo, and Twitch-explicit, 0.001 for Facebook-100, and 0.0002 for Elliptic.
 }

\textbf{Epoch.} {We follow the epochs used in EERM work and train to the end, which is 200 epochs for all datasets except for  500 epochs for OGB-Arxiv.}

\textbf{Hyperparameter Settings of $r$.}
{According to the hyperparameter studies in Sec.~\ref{para:hp}, we regard $r=0.5$ as a reasonable choice and uniformly set it for all node classification experiments with our GSINA for simplicity.}

\textbf{Performances Analysis.}
% Fig.~\ref{fig:node1},~\ref{fig:node2},~\ref{fig:node3} 
Fig.~\ref{fig:node_ood}
report the generalization performance for the distribution shifts of `Artificial Transformation', `Cross-Domain Transfers', and `Temporal Evolution', respectively. Under most testing scenarios,  GSINA  outperforms ERM  for node classification. It achieves better results than EERM on Cora, Amazon-Photo,  Elliptic, and comparable results to EERM on Twitch-explicit, Facebook-100, and OGB-Arxiv. GSINA achieves an about 20\% improvement for classification accuracy on Cora. 

\begin{table*}[tb!]
\centering
\caption{Dataset statistics of the EERM-setting benchmarks used in experiments shown in  Fig.~\ref{fig:node_ood}}
\label{tab:eerm-data}
\footnotesize
	%%\vspace{-5pt}
    \resizebox{\textwidth}{!}{
\begin{threeparttable}
\begin{tabular}{@{}cccccccc@{}}
\toprule
Datasets &    Distribution Shift & \#Nodes & \#Edges & \#Classes & Train/Val/Test Split & Metric & Aligned with \\
\midrule
 Cora    &   & 2,703 & 5,278 & 10 & by graphs & Accuracy & \cite{cora-data} \\
 Amazon-Photo   &   \multirow{-2}{*}{ Artificial Transformation}  & 7,650 & 119,081 & 10 & by graphs & Accuracy & \cite{amazon-data} \\
 Twitch-explicit    &    & 1,912 - 9,498 & 31,299 - 153,138 & 2 & by graphs & ROC-AUC & \cite{twitch-data} \\
 Facebook-100    &  \multirow{-2}{*}{Cross-Domain Transfers}  & 769 - 41,536 & 16,656 - 1,590,655 & 2 & by graphs & Accuracy & \cite{facebook100-data} \\
 Elliptic   &    & 203,769 & 234,355 & 2 & by time & F1 Score & \cite{EvolveGCN}\tnote{1} \\
 OGB-Arxiv &  \multirow{-2}{*}{Temporal Evolution} & 169,343 & 1,166,243 & 40 & by time & Accuracy & \cite{hu2020ogb} \\
\bottomrule
\end{tabular}
 \begin{tablenotes} 
         \footnotesize  
        \item[1] The original dataset is provided at \url{https://www.kaggle.com/ellipticco/elliptic-data-set}.  
       \end{tablenotes}
\end{threeparttable}
}
% %%\vspace{-6pt}
\end{table*}

\subsection{Ablation Studies on both Graph and Node Tasks}\label{sec:abl}
We provide ablation studies on both graph and node classification tasks. For graph level classification, Table~\ref{tab:abl} reports detailed experiment results on SPMotif datasets in the GSAT benchmark, the ablation versions of our GSINA are without (w/o) the Gumbel noise, Node Attention, or both (denoted as G \& N), with the same $r$ as the full GSINA (note that it is not necessary the optimal $r$ for these ablation versions). As the Node Attention is not required for node classification tasks, the ablation studies in Fig.~\ref{fig:node_ood} only provide the version without the Gumbel noise.

From these results, specific observations can be made. First, removing Gumbel noise consistently degrades performance across both graph-level and node-level tasks. In Table~\ref{tab:abl} at $b=0.5$, the accuracy of GIN+ours drops from 55.16 to 48.27, and that of PNA+ours from 76.39 to 69.95. Similarly, in Fig.~\ref{fig:node_ood} (a), the accuracy on Cora significantly decreases at transformation T4; this drop is also present in Elliptic T9 and OGB-Arxiv. In addition to mean performance, we observe that the performance variance becomes higher after removing Gumbel noise, as indicated by the wider shaded regions in Figs.~6–8. This demonstrates that the Gumbel trick not only improves average generalization performance but also stabilizes training dynamics. We attribute this to the stochasticity introduced in the Sinkhorn optimization, which prevents deterministic attention distributions from overfitting to spurious subgraphs and encourages better exploration of diverse invariant structures. Notably, the benefit of Gumbel is more pronounced under lower $b$ values (e.g., $b=0.5$), where spurious signals are weaker and discriminative invariant features are harder to learn—this highlights its importance for robustness in weak-signal scenarios.

Node Attention, on the other hand, plays a crucial role specifically for graph-level classification tasks. By assigning attention weights to nodes, it allows the model to selectively emphasize task-relevant parts of the subgraph. In Table~\ref{tab:abl}, at $b=0.7$, removing Node Attention causes PNA+ours to drop from 73.96 to 58.70, and GIN+ours from 56.83 to 54.63. The impact is particularly notable when the underlying task-relevant structure is sparse and localized (as is often the case in motif-based datasets), requiring the model to reason over node importance beyond simple topological proximity. We also find that the effectiveness of Node Attention decreases slightly as $b$ increases (e.g., $b=0.9$), which we interpret as the model needing to rely more on global graph context when the spurious subgraphs dominate.

Importantly, the variant that removes both Gumbel noise and Node Attention performs the worst across nearly all settings. For instance, at $b=0.5$, GIN+ours drops to 46.28, and PNA+ours to 71.75, clearly indicating that the two components are not only individually beneficial but also complementary. Gumbel noise introduces necessary randomness for better subgraph candidate exploration, while Node Attention ensures that once a subgraph is extracted, the model can still identify the most informative regions within it. Their combination enables GSINA to learn subgraphs that are simultaneously stable, discriminative, and robust to distribution shifts.

We also compare our full model against GSAT-based baselines with the same GIN and PNA backbones. While GSAT adopts a hard top-$k$ subgraph selection strategy, it lacks differentiability and soft control, often resulting in truncated or suboptimal subgraphs. In contrast, GSINA's differentiable and softly constrained mechanism produces better representations. For example, at $b=0.5$, PNA+GSAT achieves 68.74, whereas PNA+GSINA achieves 76.39. This gap remains consistent across varying $b$, validating the advantage of our soft attention with separability and softness constraints, which together form a more flexible and optimizable framework for invariant subgraph learning.

% From the ablation studies, we observe the performance degradations for graph and node level tasks when learning GSINA without Gumbel noise, Node Attention, or both, demonstrating the effectiveness of the proposed components of GSINA. In the graph classification setting, our degenerated versions i.e. w/o Gumbel, w/o NodeAttn, w/o G\&N show notable performance drop compared with our complete version namely GIN+GSINA or PNA+GSINA, across different values of the hyperparamter $b$ for dataset generation. Moreover, in node classification tasks, the ablation versions provide higher variances as illustrated by the darkness area along the curves, in the datasets of Cora, Amazon-Photo, and OGB-Arxiv, suggesting that the Gumbel trick could stabilize GSINA learning.

\begin{table}[t!]%{R}{0.5\columnwidth}
% %%\vspace{-20pt}
\caption{Ablation studies for the graph level OOD task (classification AUC) w.r.t. the bias hyperparameter $b$ for graph generation on the Spurious-motif dataset.}
%%\vspace{-5pt}
\begin{center}
% \begin{small}
% \resizebox{0.5\columnwidth}{!}{
% \begin{sc}
\begin{tabular}{lccc}
\toprule
  % & \multirow{2}{*}{MolHiv (AUC)} & \multirow{2}{*}{Graph-SST2} & \multirow{2}{*}{MNIST-75sp}
% & \multicolumn{3}{c}{Spurious-motif}                  \\
  % &                    &                    &  
  Methods & $b=0.5$                  & $b=0.7$                  & $b=0.9$                  \\
\midrule
% %\midrule
GIN~\cite{xu2018powerful}+GSAT
% & $76.47{\scriptstyle \pm1.53}$
% & $82.95{\scriptstyle \pm0.58}$
% & $96.24{\scriptstyle \pm0.17}$
& $52.74{\scriptstyle \pm4.08}$ & $49.12{\scriptstyle \pm3.29}$ & $44.22{\scriptstyle \pm5.57 }$\\
 GIN~\cite{xu2018powerful}+GSINA (ours) & 
% $-$ & 
% $\mathbf{83.66}{\scriptstyle \pm0.37}$ & % r = 0.2
% $\mathbf{96.73}{\scriptstyle \pm0.16}$ & % r = 0.5
$\mathbf{55.16}{\scriptstyle \pm5.69}$ & % r=0.6
$\mathbf{56.83}{\scriptstyle \pm6.32}$ &  %r=0.6
${49.86}{\scriptstyle \pm6.10}$\\ % r=0.4
 w/o Gumbel 
%  & 
% $-$ & 
% $83.45{\scriptstyle \pm0.42}$ 
% & 
% $96.87{\scriptstyle \pm0.22}$ 
& $48.27{\scriptstyle \pm4.80}$ & $45.25{\scriptstyle \pm7.15}$ & $\mathbf{50.28}{\scriptstyle \pm2.83}$  \\ 

 w/o NodeAttn 
%  & 
% $-$ 
% & 
% $83.41{\scriptstyle \pm1.09}$ 
% & $96.47{\scriptstyle \pm0.46}$ 
& $47.34{\scriptstyle \pm7.99}$ & $54.63{\scriptstyle \pm6.99}$ & $48.41{\scriptstyle \pm1.16}$  \\ 
 w/o G\&N 
%  & $-$ 
%  & 
% $84.07{\scriptstyle \pm0.38}$ 
% & $96.87{\scriptstyle \pm0.28}$ 
& $46.28{\scriptstyle \pm5.67}$ & $45.40{\scriptstyle \pm3.22}$ & $44.44{\scriptstyle \pm6.56}$  \\ 
\midrule
% PNA & $78.91{\scriptstyle \pm1.04}$ & $79.87{\scriptstyle \pm1.02}$ & $87.20{\scriptstyle \pm5.61}$ & $68.15{\scriptstyle \pm2.39}$ & $66.35{\scriptstyle \pm3.34}$ & $61.40{\scriptstyle \pm3.56} $ \\
PNA~\cite{corso2020principal}+GSAT 
% & $80.24{\scriptstyle \pm0.73}$ 
% & $80.92{\scriptstyle \pm0.66}$ 
% & $93.96{\scriptstyle \pm0.92}$ 
& $68.74{\scriptstyle \pm2.24}$ & $64.38{\scriptstyle \pm3.20}$ & $57.01{\scriptstyle \pm2.95}$  \\
PNA~\cite{corso2020principal}+GSINA (ours) & 
% $\mathbf{80.41}{\scriptstyle \pm1.15}$ & % r=0.9
% $\mathbf{82.18}{\scriptstyle \pm1.01}$ & % r = 0.8
% $\mathbf{95.48}{\scriptstyle \pm0.37}$ & % r = 0.6
$\mathbf{76.39}{\scriptstyle \pm1.85}$ & % r=0.1
$\mathbf{73.96}{\scriptstyle \pm2.87}$ &  %r=0.3
$\mathbf{62.51}{\scriptstyle \pm5.86}$\\ % r=0.5
w/o Gumbel 
% & 
% $-$ 
% & 
% $82.12{\scriptstyle \pm0.98}$ 
% & $94.59{\scriptstyle \pm0.98}$ 
& $69.95{\scriptstyle \pm2.76}$ & $69.67{\scriptstyle \pm3.44}$ & $62.14{\scriptstyle \pm4.64}$  \\ 

 w/o NodeAttn 
%  & 
% $-$ 
% & 
% $81.42{\scriptstyle \pm0.77}$ 
% & $95.51{\scriptstyle \pm0.18}$ 
& $71.60{\scriptstyle \pm1.89}$ & $58.70{\scriptstyle \pm3.82}$ & $58.20{\scriptstyle \pm2.70}$  \\ 

 w/o G\&N 
%  & 
% $-$ 
% & 
% $82.14{\scriptstyle \pm1.05}$ 
% & $93.98{\scriptstyle \pm1.38}$ 
& $71.75{\scriptstyle \pm2.42}$ & $67.50{\scriptstyle \pm4.51}$ & $61.34{\scriptstyle \pm1.72}$  \\
\bottomrule
% \label{table:Generalization}
\end{tabular}
% \end{sc}
% }
% \end{small}
\end{center}
\label{tab:abl}
     %%\vspace{-20pt}
% \end{table*}
\end{table}

\begin{table*}[tb!]
\caption{Training time comparison on the benchmark used in CIGA~\cite{chen2022learning}. We record the training time in terms of s/epoch.}\label{app:time}
\centering
\resizebox{\textwidth}{!}{
\begin{tabular}{lllllllllll}
\toprule
        Methods    & Spmotif                            & Graph-SST5                         & Twitter                            & Drug-Assay                         & Drug-Sca                           & Drug-Size                           & TU-NCI1                           & TU-NCI109                         & TU-PROT                           & TU-DD                             \\ \midrule
ERM~\cite{wu2022handling}         & $5.80 { \scriptstyle \pm 0.22 } $  & $3.99 { \scriptstyle \pm 0.10 } $  & $2.12 { \scriptstyle \pm 0.29 } $  & $9.77 { \scriptstyle \pm 2.12 } $  & $6.70 { \scriptstyle \pm 0.66 } $  & $10.81 { \scriptstyle \pm 1.10 } $  & $1.26 { \scriptstyle \pm 0.07 } $ & $1.22 { \scriptstyle \pm 0.08 } $ & $0.40 { \scriptstyle \pm 0.08 } $ & $0.37 { \scriptstyle \pm 0.09 } $ \\ 
GSAT~\cite{miao2022interpretable}        & $11.08 { \scriptstyle \pm 0.29 } $ & $7.74 { \scriptstyle \pm 0.08 } $  & $4.13 { \scriptstyle \pm 0.14 } $  & $17.03 { \scriptstyle \pm 0.66 } $ & $10.85 { \scriptstyle \pm 0.60 } $ & $17.95 { \scriptstyle \pm 1.07 } $  & $2.45 { \scriptstyle \pm 0.10 } $ & $2.39 { \scriptstyle \pm 0.08 } $ & $0.71 { \scriptstyle \pm 0.08 } $ & $0.72 { \scriptstyle \pm 0.09 } $ \\ 
GSINA-macro & $14.80 { \scriptstyle \pm 0.32 } $ & $9.50 { \scriptstyle \pm 0.23 } $  & $5.21 { \scriptstyle \pm 0.13 } $  & $19.83 { \scriptstyle \pm 1.24 } $ & $12.68 { \scriptstyle \pm 0.67 } $ & $23.15 { \scriptstyle \pm 1.54 } $  & $3.14 { \scriptstyle \pm 0.12 } $ & $3.00 { \scriptstyle \pm 0.09 } $ & $0.89 { \scriptstyle \pm 0.07 } $ & $1.04 { \scriptstyle \pm 0.06 } $ \\ 
CIGA~\cite{chen2022learning}        & $21.66 { \scriptstyle \pm 0.91 } $ & $14.34 { \scriptstyle \pm 0.25 } $ & $7.96 { \scriptstyle \pm 0.21 } $  & $54.83 { \scriptstyle \pm 1.22 } $ & $33.74 { \scriptstyle \pm 0.62 } $ & $58.45 { \scriptstyle \pm 0.92 } $  & $5.01 { \scriptstyle \pm 0.13 } $ & $4.48 { \scriptstyle \pm 0.05 } $ & $1.19 { \scriptstyle \pm 0.08 } $ & $1.36 { \scriptstyle \pm 0.11 } $ \\ 
GSINA-micro & $34.58 { \scriptstyle \pm 0.63 } $ & $24.52 { \scriptstyle \pm 0.37 } $ & $13.04 { \scriptstyle \pm 0.21 } $ & $97.43 { \scriptstyle \pm 3.12 } $ & $59.47 { \scriptstyle \pm 0.69 } $ & $101.28 { \scriptstyle \pm 0.75 } $ & $7.43 { \scriptstyle \pm 0.11 } $ & $7.37 { \scriptstyle \pm 0.13 } $ & $1.93 { \scriptstyle \pm 0.09 } $ & $2.14 { \scriptstyle \pm 0.11 } $ \\ 
\bottomrule
\end{tabular}
}
% %%\vspace{-15}
\end{table*}
% (we always select the better one for experiments)
\subsection{Runtime and Complexity Analysis} 
We evaluate GSINA in two settings for subgraph extraction in a batch. The first (denoted as `micro') setting is to compute soft top-$r$ for each graph in the batch, which is slower, and we use it on Spmotif, TU-PROT, TU-DD, Graph-SST2, MNIST-75sp, OGBG-Molbbbp, OGBG-Molclintox, OGBG-Moltox21, OGBG-Molsider. The other (denoted as `macro') setting is to compute soft top-$r$ only once for the entire batch (which can be considered as a large graph composed of several smaller graphs), being faster. We apply it on Graph-SST5, Twitter, Drug-Assay, Drug-Sca, Drug-Size, TU-NCI1, TU-NCI109, OGBG-Molhiv, OGBG-Molbace. The macro setting computes the top-$r$ for the whole graph and relaxes the constraint of top-$r$ on each individual graph, which is useful when the distribution shifts are complicated, and in that case, it is difficult to find a uniform reasonable $r$ for each graph. Therefore, the macro v.s. micro model selection in GSINA is highly dependent on the characteristics of the dataset, and we select the one with better performance for each dataset. On the other hand, CIGA uses hard top-$k$ selection for subgraph extraction for each graph (similar to the micro setting) in the batch. We have implemented GSAT (without hyperparameter fine-tuning) on the CIGA benchmark for training time comparison, GSAT is similar to the macro setting to an extent, and GSAT models the probability $p$ of each edge belonging to the invariant part, while it does not consider how many graphs are in the batch.

 We conduct experiments regarding model training time on all datasets of the CIGA benchmark. Table~\ref{app:time} reports training time comparison on CIGA. Note that ERM is always the fastest, and our second setting (GSINA-macro) is often much faster than CIGA and a bit slower than GSAT due to iterative computations by Eq.~\ref{eq:sinkhorn0},~\ref{eq:sinkhorn1},~\ref{eq:sinkhorn2}; while the first setting (GSINA-micro), also due to iterative computations in subgraph extraction, is slower than CIGA's hard top-k selection. However, since we can set the number of iterations to a relatively small value (uniformly set to 10 in this paper), GSINA does not introduce a significant overhead and is always within two times of CIGA's run time.

% %%\vspace{-2pt}
\begin{table}[t!]
% %%\vspace{-10pt}
\caption{Interpretation Performance (AUC) with the varying bias hyperparameter $b$ used for data generation.}
\begin{center}
\begin{tabular}{lccc}
\toprule
Methods
  & \multicolumn{3}{c}{Spurious-motif}                                        \\
                 % &  
                 & $b=0.5$                  & $b=0.7$                  & $b=0.9$                  \\
\midrule
GNNExplainer~\cite{ying2019gnnexplainer}
% & $59.01{\scriptstyle \pm2.04}$
& $62.62{\scriptstyle \pm1.35}$ & $62.25{\scriptstyle \pm3.61}$ & $58.86{\scriptstyle \pm1.93}$ \\
PGExplainer~\cite{luo2020parameterized}
% & $69.34{\scriptstyle \pm4.32}$ 
& $69.54{\scriptstyle \pm5.64}$ & $72.33{\scriptstyle \pm9.18}$ & ${72.34}{\scriptstyle \pm2.91}$ \\
GraphMask~\cite{schlichtkrull2021interpreting}   
% & $\underline{73.10}{\scriptstyle \pm6.41}$ 
& $72.06{\scriptstyle \pm5.58}$ & $73.06{\scriptstyle \pm4.91}$ & $66.68{\scriptstyle \pm6.96}$ \\
GIB~\cite{yu2020graph}         
% & $51.20{\scriptstyle \pm5.12}$
& $57.29{\scriptstyle \pm14.35}$ & $62.89{\scriptstyle \pm15.59}$ & $47.29{\scriptstyle \pm13.39}$ \\
DIR~\cite{wu2022discovering}      
% & $32.35{\scriptstyle \pm9.39}$
& ${78.15}{\scriptstyle \pm1.32}$ & ${77.68}{\scriptstyle \pm1.22}$ & $49.08{\scriptstyle \pm3.66}$ \\
\midrule
GIN~\cite{xu2018powerful}+GSAT     
% & $\mathbf{83.36}{\scriptstyle \pm1.02}$
& $\underline{78.45}{\scriptstyle \pm3.12}$ & $74.07{\scriptstyle \pm5.28}$ & $71.97{\scriptstyle \pm4.41}$ \\
GIN~\cite{xu2018powerful}+GSINA (ours) 
% & $55.60 {\scriptstyle \pm3.39}$ 
& $65.13 {\scriptstyle \pm10.00}$ & $60.78 {\scriptstyle \pm8.09}$ & $55.86 {\scriptstyle \pm3.56}$\\
\midrule
PNA~\cite{corso2020principal}+GSAT   
% & $\mathbf{84.68}{\scriptstyle \pm1.06}$ 
& $\mathbf{83.34}{\scriptstyle \pm2.17}$ & $\mathbf{86.94}{\scriptstyle \pm4.05}$ & $\mathbf{88.66}{\scriptstyle \pm2.44}$ \\
PNA~\cite{corso2020principal}+GSINA (ours) 
% & $61.16{\scriptstyle \pm2.76}$
& $75.66{\scriptstyle \pm1.56}$ & $\underline{80.47}{\scriptstyle \pm1.06}$ & $\underline{80.10}{\scriptstyle \pm2.04}$\\
\bottomrule
\label{tab:xauc}
\end{tabular}
% \end{sc}
% \end{small}
\end{center}
\end{table}

\begin{table}[t!]
\caption{Interpretation precision@5 of with the GNN backbone `SPMotifNet' in DIR~\cite{wu2022discovering}.}
%%\vspace{-10pt}
\begin{center}
% \begin{small}
% \begin{sc}
\begin{tabular}{lccc}
\toprule
  & \multicolumn{3}{c}{Spurious-motif}                   \\
  Methods& $b = 0.5$          & $b=0.7$           & $b=0.9$           \\
\midrule
GNNExplainer~\cite{ying2019gnnexplainer} & $0.203\scriptstyle{\pm0.019}$ & $0.167\scriptstyle{\pm0.039}$ & $0.066\scriptstyle{\pm0.007}$ \\
DIR~\cite{wu2022discovering} & $0.255\scriptstyle{\pm0.016}$ & $0.247\scriptstyle{\pm0.012}$ & $0.192\scriptstyle{\pm0.044}$ \\
GSAT~\cite{miao2022interpretable} & $\mathbf{0.519}\scriptstyle{\pm0.022}$ & $\mathbf{0.503}\scriptstyle{\pm0.034}$ & $\underline{0.416}\scriptstyle{\pm0.081}$ \\
GSINA (ours) &$\underline{0.419} \scriptstyle{\pm0.030}$& $ \underline{0.401 }\scriptstyle{\pm 0.046}$ & $\mathbf{0.429} \scriptstyle{\pm 0.040}$\\
\bottomrule
\label{tab:prec5}
\end{tabular}
% \end{sc}
% \end{small}
\end{center}
    %%\vspace{-25pt}
% %%\vspace{-9mm}
\end{table}

\subsection{Further Discussion}\label{app:int}
% %%\vspace{-2pt}
We now give further discussion on the interpretability and other properties of the extracted subgraph.
\subsubsection{Subgraph Interpretability}
In addition to the generalizability detailedly discussed above, here we discuss the interpretability of GSINA regarding the accuracy of the invariant subgraph extraction. The interpretability is evaluated based on SPMotif datasets, which have labeled ground truths of explanation subgraphs (a.k.a invariant subgraphs $G_S$). The interpretability evaluation metrics follow the previous studies DIR~\cite{wu2022discovering} and GSAT~\cite{miao2022interpretable} for the performance of explanation subgraph recognition, which is a problem of binary classification for each edge. Following DIR and GSAT, we perform edge binary classifications, for metrics, we evaluate ROC-AUC (with GIN and PNA backbones, in line with the setting of GSAT) and precision@5 (with the GNN backbone `SPMotifNet' used in DIR~\cite{wu2022discovering}).

As shown in Table~\ref{tab:xauc} and Table~\ref{tab:prec5}, GSINA (using PNA and SPMotifNet as backbones) mostly outperform interpretable GNNs DIR and GIB as well as other post-hoc~\cite{miao2022interpretable} GNN explainers, showing the inherent interpretability to extract invariant subgraphs. However, GSINA is inferior to GSAT in interpretability. In our analysis, it is due to the innate characteristics of top-$k$ based methods: intuitively, for a `hard' top-$k$, the output value is binary (0 / 1), meaning an item belongs to top-$k$ or not. In another word, `hard' top-$k$ does not consider the relative ranking of items at all. While the prevalent metrics for binary classification (e.g. AUC, precision@5) always do. Hence, it is not natural to evaluate the subgraph recognition performance of top-$k$ based methods. However, GSAT does not restrict edge predictions to (approximated) binary like top-$k$ methods, so that it achieves more flexible edge attention weights and better performance in interpretability. 

In GSINA, the use of a soft top-$r$ operation imposes stronger constraints on the attention value distribution than GSAT. The choice of subgraph ratio $r$ can therefore influence the attention distribution, whereas GSAT is not directly affected by such a parameter. Overall, GSINA is designed to balance interpretability and generalization. Although it may yield slightly lower scores on pure interpretability metrics, it achieves substantially stronger OOD generalization, aligning with the core objective of this study.

% In contrast, due to the utilization of soft top-$r$ operation in GSINA, it places more constraints on the graph attention value distribution compared to GSAT. 
% Note that the choice of the subgraph ratio $r$ in GSINA could significantly impact the attention distribution, whereas GSAT is not. To sum up, the design of GSAT, as mentioned above, is a tradeoff. It excels in interpretability, but it falls short in effectively filtering out variant parts (as described in Sec.~\ref{sec:intro} and Fig.~\ref{fig:demo}). This leads to GSAT may have less ability to make predictions using subgraphs compared to GSINA.

% as also empirically observed in our experiments.

\begin{figure}[tb!]
     \centering
     \begin{subfigure}[b]{0.48\textwidth}
         \centering
    \includegraphics[width=\textwidth]{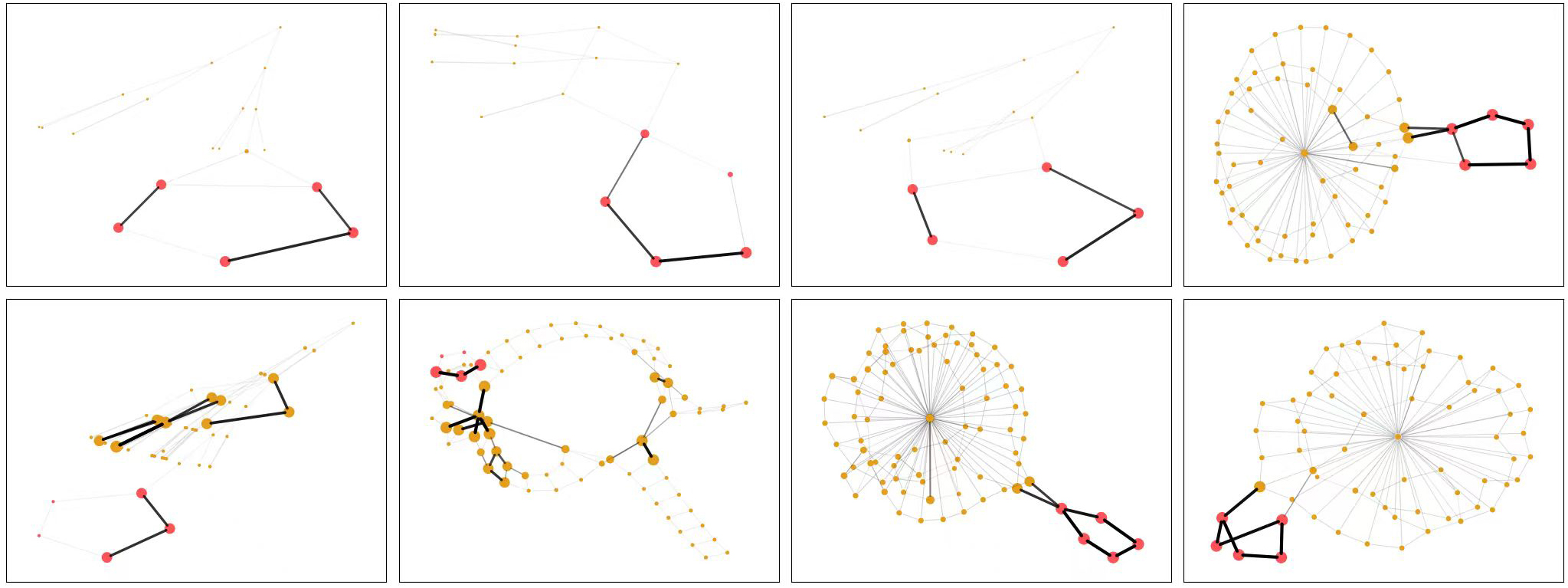}
         \caption{SPMotif-0.5, label  0 (motif is cycle)}
         % \label{fig:y equals x}
     \end{subfigure}
     \begin{subfigure}[b]{0.48\textwidth}
         \centering
\includegraphics[width=\textwidth]{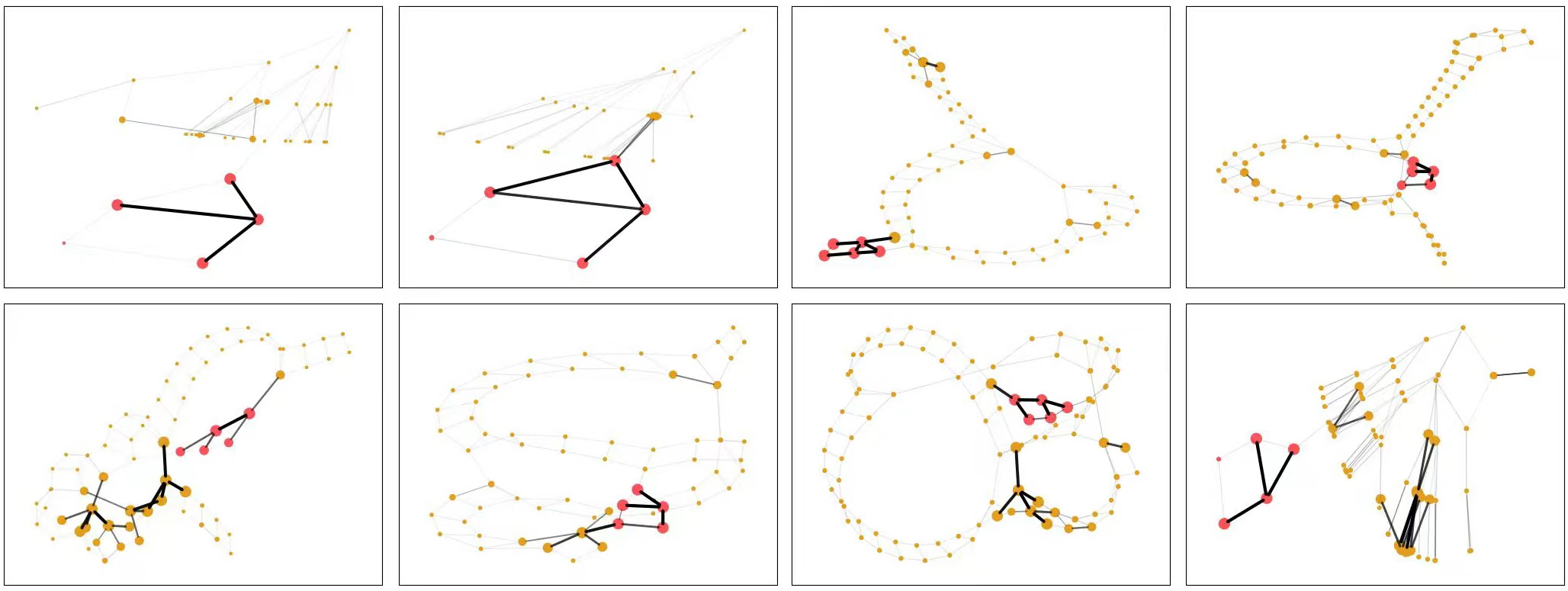}
         \caption{SPMotif-0.5, label  1 (motif is house)}
         % \label{fig:y equals x}
     \end{subfigure}
     \begin{subfigure}[b]{0.48\textwidth}
        \centering
     \includegraphics[width=\textwidth]{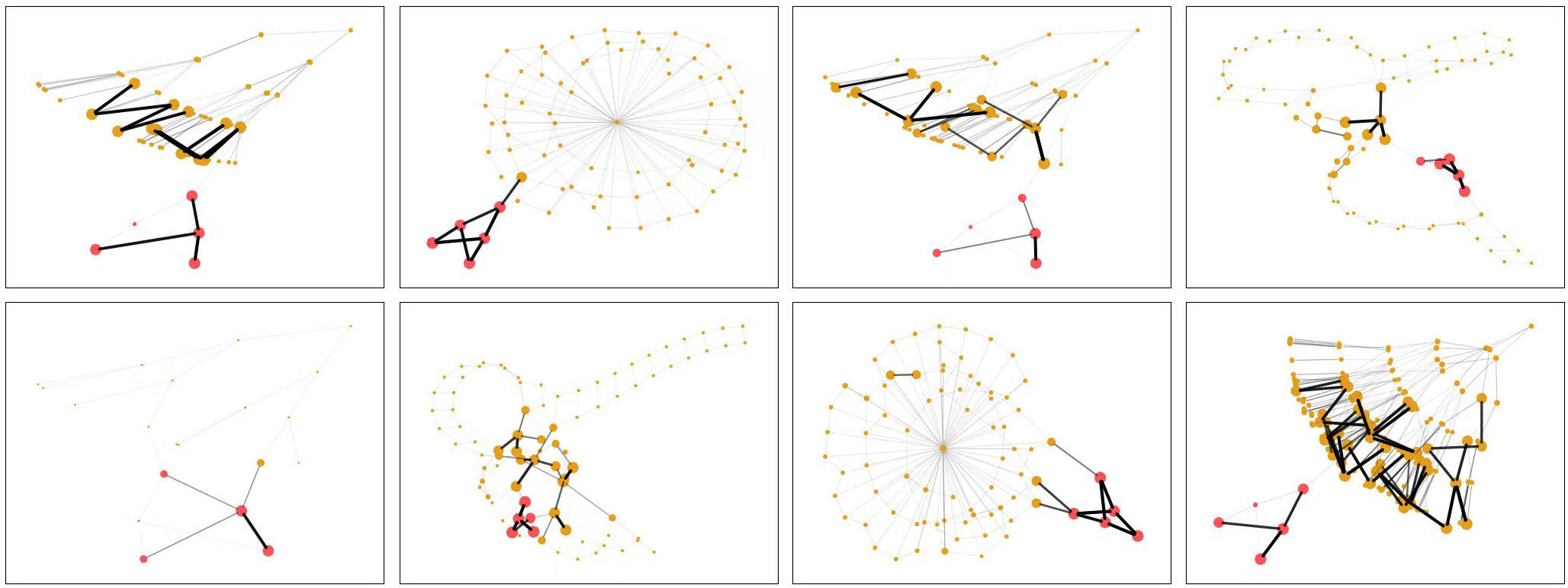}
         \caption{SPMotif-0.5, label  2 (motif is crane)}
         % \label{fig:y equals x}
     \end{subfigure}
     \caption{Visualizing the extracted subgraphs in SPMotif-0.5, i.e. the generation hyperparameter $b$ is set to 0.5.}
     \label{fig:vis_SPMotif}
\end{figure}

\begin{figure}[tb!]
     \begin{subfigure}[b]{0.48\textwidth}
         \centering
         \includegraphics[width=\textwidth]{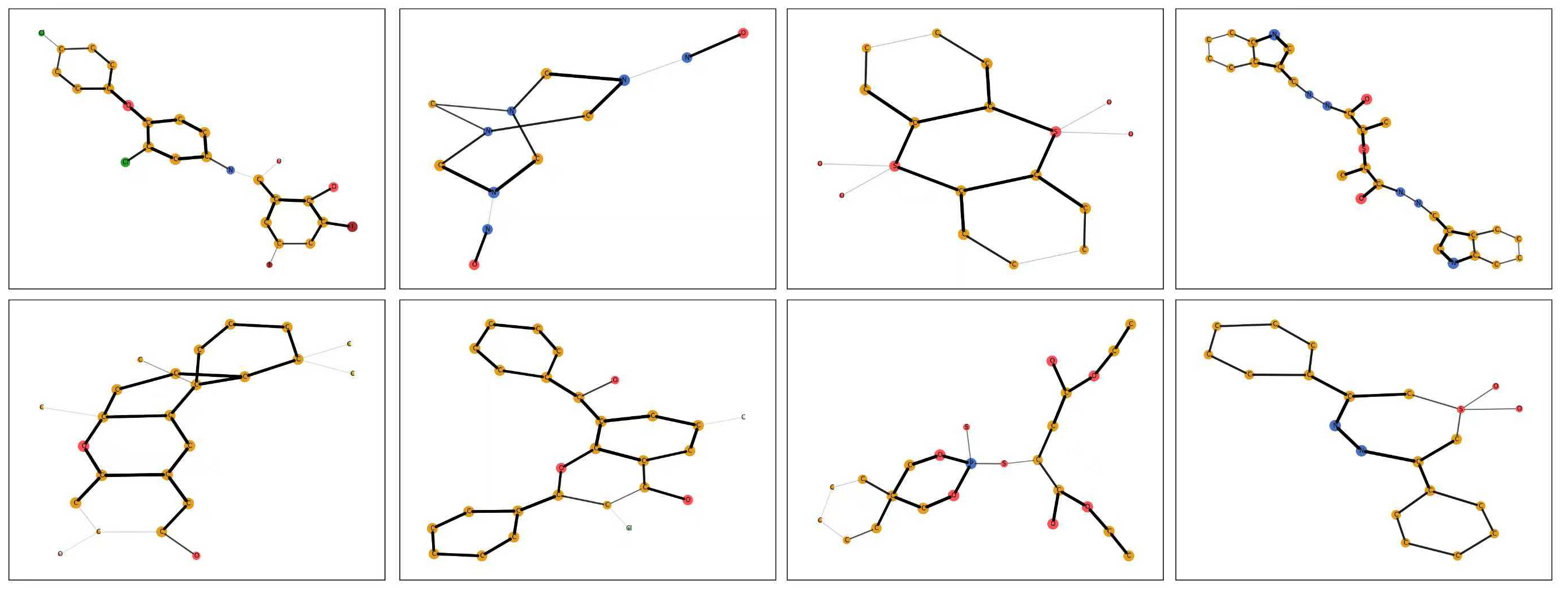}
         \caption{OGBG-Molhiv, label 0 (not inhibits HIV virus replication)}
         % \label{fig:y equals x}
     \end{subfigure}

     \begin{subfigure}[b]{0.48\textwidth}
         \centering
         \includegraphics[width=\textwidth]{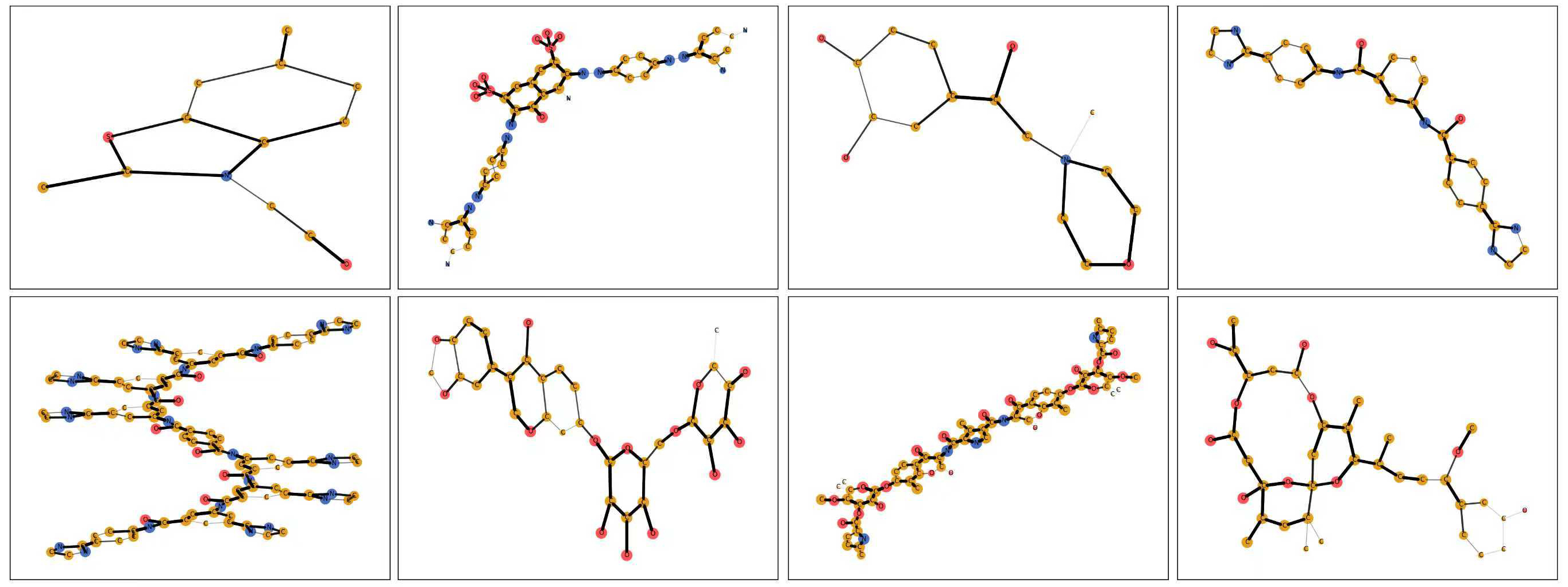}
         \caption{OGBG-Molhiv, label 1 (inhibits HIV virus replication)}
         % \label{fig:y equals x}
     \end{subfigure}
\caption{Visualizing extracted subgraphs in OGBG-Molhiv.}\label{fig:vis_OGBG}
\end{figure}

\subsubsection{Subgraph Property and Visualization}
Recall that we take three properties in this paper: separability, softness, and differentiability, into consideration to improve the invariant subgraph extraction for GIL. However, it is definitely not the ultimate solution to GIL subgraph extraction. There might be other ad-hoc considerations of how the extracted invariant subgraph should be, e.g. subgraph connectivity and completeness, as it might be expected that the invariant regions should represent coherent structures and patterns rather than disjoint disconnected components.
For such considerations, we provide elementary solutions and discuss the difficulties of the problems.

%the crucial subgraphs might be either connected or not. For example, a molecule may have multiple disconnected functional groups playing a key role. Connectivity
%we believe that it depends on the characteristics of the data. T
For subgraph connectivity, it is not explicitly considered in our GSINA design. One may resort adding a penalty loss (like the connectivity loss in GIB~\cite{yu2020graph}) to our learning objective in Sec.~\ref{sec:learning_obj} to enhance connectivity. However, the subgraph is latent and inferred by the GNN from data, and the understanding of data by the model does not necessarily have to be consistent with that of humans. On the other hand, what humans consider a `complete subgraph' may still have redundant parts in the model's learning process. It is difficult for even experts to determine definitively which nodes or edges should be included in a `complete invariant subgraph'. For the other property of completeness, since many datasets may lack ground-truth information about the invariant subgraph, it is also challenging to define a metric or penalty for subgraph completeness. Currently, even without a guarantee of subgraph connectivity or completeness, GSINA could still improve GIL performances based on its extracted subgraphs. We leave a more in-depth design and modification for future work.

%\subsection{Visualization}\label{sec:vis}
In Fig.~\ref{fig:vis_SPMotif} and Fig.~\ref{fig:vis_OGBG}, we visualize the subgraph extractions on datasets in the GSAT benchmark. Specifically, we visualize the original edge and node attention values given by our GSINA, and do not perform the scaling tricks: first normalization, then multiplying edges' numerical values repeatedly (10 times in GSAT) to improve discrimination, as also applied in the visualizations of GSAT and CIGA.

\section{Conclusion}\label{sec:con}
We have developed Graph Sinkhorn Attention (GSINA), a general invariance optimization framework for Graph Invariant Learning (GIL) to improve the generalization for both graph and node level tasks by extracting the sparse, soft, and differentiable invariant subgraphs in the manner of graph attention. Theoretical study is given for the convergence behavior of our alternating iterative algorithm. Extensive experimental results across different settings and benchmarks have shown the superiority of GSINA against the state-of-the-arts on both graph and node level GIL tasks.

For outlook, we expect the community of graph learning would expand the frontier to more complex tasks e.g. combinatorial optimization~\cite{YanIJCAI20}, which has been an emerging area for further efforts whereby various constraints need be carefully handled and the technique used in this paper namely Skinhhorn iteration has underlying connection to specific problems e.g. graph matching~\cite{WangPAMI22}, can be a powerful tool for more broad problems~\cite{ShiICLR25}. Another promising area is developing the quantum counterpart of learning with graphs~\cite{YeICML23} and quantum inspired graph methods~\cite{TangNIPS22}.

% In this work, we introduced Graph Sinkhorn Attention (GSINA), a comprehensive invariance optimization framework for Graph Invariant Learning (GIL) to improve the generalization for both graph and node level tasks by extracting sparse, soft, and differentiable invariant subgraphs in the manner of graph attention. 
% Extensive experimental results proved the superiority of GSINA over state-of-the-art approaches for both graph and node level GIL tasks.

\section*{Acknowledgment}
This work was in part supported by NSFC (92370201, 62222607) and Ant Group. %The authors are thankful to the constructive suggestions provided by the reviewers and associate editor.
%and also to Prof. Liangliang Shi, Dr. Tianming Li and Dr. Haiyang Wang for their valuable suggestions to improve the paper.

\bibliographystyle{IEEEtran}  
\bibliography{ref}
% \bibliographystyle{abbrv}

% \vspace{-20pt}
\begin{IEEEbiography}
[{\includegraphics[width=1in,height=1.25in,clip,keepaspectratio]{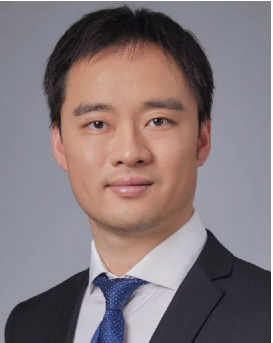}}] 
{Junchi Yan} (S'10-M'11-SM'21) is currently a Professor with School of Artificial Intelligence (also affiliated with School of Computer Science), Shanghai Jiao Tong University (SJTU), Shanghai, China. Before he joined SJTU in 2018, he was a (Senior) Research Staff Member with IBM Research (2011-2018). He was also an affiliated Researcher with AWS AI Lab (2019-2022). His research interests are machine learning and applications. He serves as the Associate Editor for IEEE TPAMI, IEEE TNNLS, IEEE TEVC, TMLR, Pattern Recognition, and Senior Editor for ACM Trans. on Probabilistic Machine Learning. He regularly serves Area Chair for CVPR, ECCV, ICML, ICLR, NeurIPS etc. He obtained his PhD degree in electrical engineering from SJTU in 2015. He received the IEEE CS AI'10 to Watch, IEEE CIS Outstanding Early Career Award, as well as IROS 2025 Best Paper Finalist, CVPR 2024 Best Paper Candidate, and ACL 2025 Outstanding Paper. He is a Fellow of IAPR, IET, and on the board of ICML and a Cover Author of IEEE Xplorer.
\end{IEEEbiography}

% \vspace{-20pt}
\begin{IEEEbiography}
[{\includegraphics[width=1in,height=1.25in,clip,keepaspectratio]{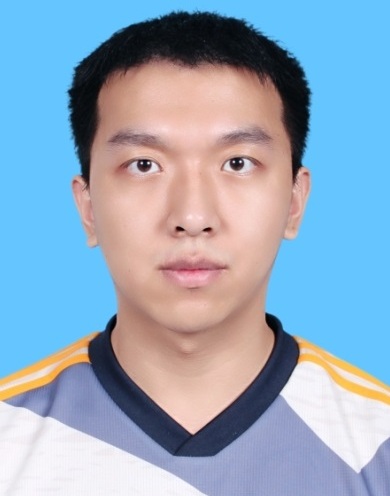}}] 
{Fangyu Ding} obtained both the master and Bachelor degrees in Department of Computer Science and Engineering, Shanghai Jiao Tong University, Shanghai, China. He currently is a PhD student with HKUST, Honkong, China. His research interests include machine learning and systems. He is now a PhD Student with Hongkong University of Science and Technology, Hongkong, China.
\end{IEEEbiography}

% \vspace{-20pt}

\begin{IEEEbiography}
[{\includegraphics[width=1in,height=1.25in,clip,keepaspectratio]{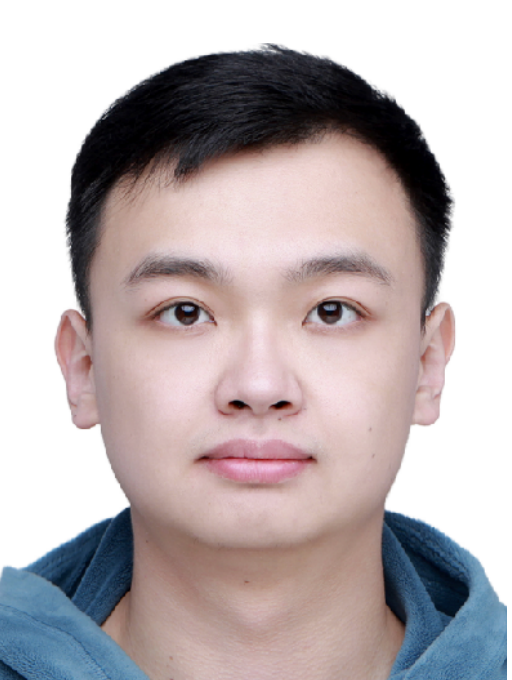}}] {Jiawei Sun} received the B.E. degree in electrical and computer engineering from Shanghai Jiao Tong University (SJTU), Shanghai, China, in 2020, and the Ph.D. degree in computer science in the same university in 2025. He is working in the field of graph neural networks, self-supervised learning, and spatio-temporal data mining. He has published first-authored papers in ICLR and Neural Networks.\end{IEEEbiography}

% \vspace{-20pt}

%\begin{IEEEbiography}
%[{\includegraphics[width=1in,height=1.25in,clip,keepaspectratio]{author/wang}}] 
%{Haiyang Wang} is a research engineer at Ant Group. Before joining Ant Group, he worked as a quantitative researcher at Huatai Security (Shanghai). He received his Ph.D. from Shanghai Jiao Tong University, specializing in Information Engineering. He completed his undergraduate studies at Shanghai Jiao Tong University. His main research interests include: spatial-temporal data mining, recommendation algorithms, time series prediction.
%\end{IEEEbiography}

\begin{IEEEbiography}[{\includegraphics[width=1in,height=1.25in,clip,keepaspectratio]{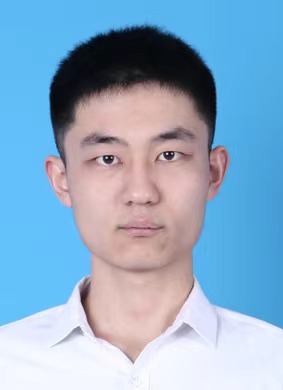}}] {Zhaoping Hu} received the M. E. degree in computer science and technology from Shanghai Jiao Tong University, China in 2024, and received his B.E. in Computer Science from Dalian University of Technology, in 2021. His research interests include spatial-temporal data mining.\end{IEEEbiography}

% \vspace{-20pt}
\begin{IEEEbiography}
[{\includegraphics[width=1in,height=1.25in,clip,keepaspectratio]{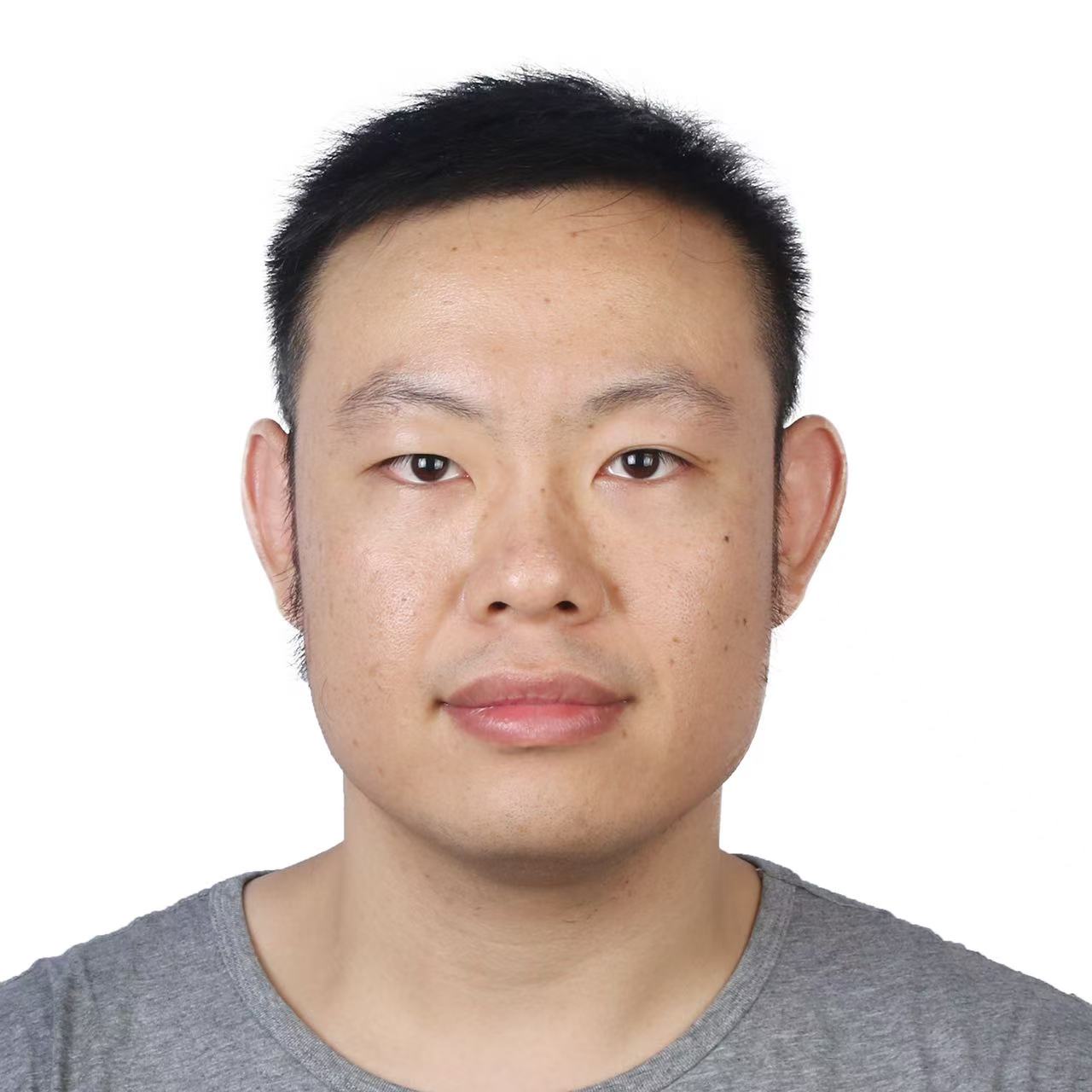}}] 
{Yunyi Zhou} received his B.S. degree from Hunan Normal University, Changsha, China in 2013 in Statistics. In 2018, he obtained his Ph.D. degree from School of Mathematical Siences, Fudan University, Shanghai, China. His research interests include time series forecasting, data mining and large language models. Currently, he is an Algorithm Expert at Ant Group.
\end{IEEEbiography}
%\vspace{-20pt}
\begin{IEEEbiography}
[{\includegraphics[width=1in,height=1.25in,clip,keepaspectratio]{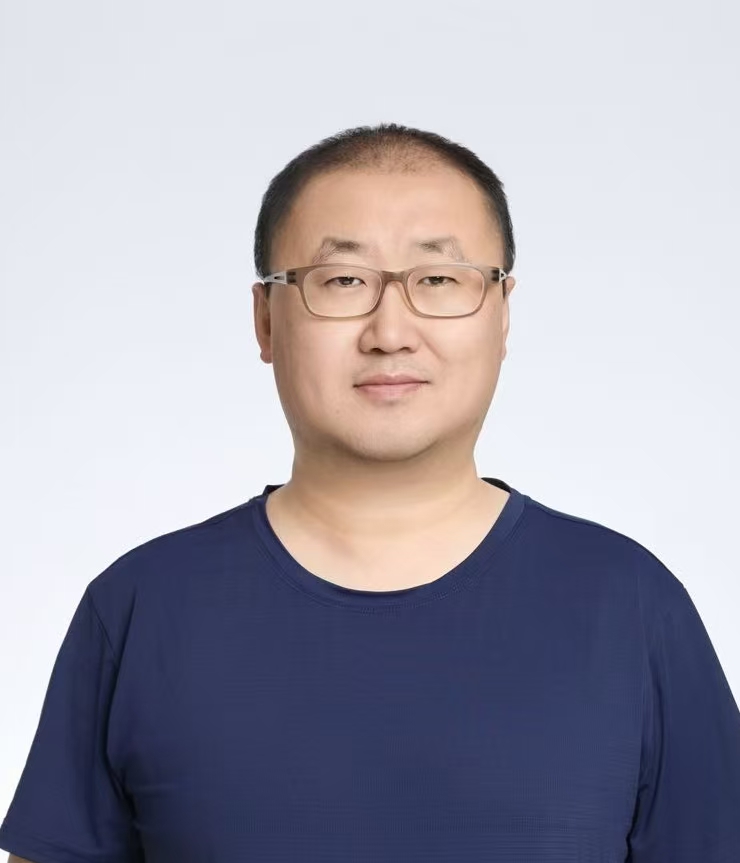}}] 
{Lei Zhu} is a Senior Algorithm Expert at Ant Group with visiting scholar experience at the University of California, San Diego (UCSD) and Newcastle University. His research interests include machine learning, large language model applications, and financial intelligence.
\end{IEEEbiography}

\end{document}